\documentclass{article}
\usepackage{ijcai25}
\usepackage{times}
\usepackage{soul}
\usepackage{url}
\usepackage[hidelinks]{hyperref}
\usepackage[utf8]{inputenc}
\usepackage[small]{caption}
\usepackage[font=footnotesize,labelfont=rm,textfont=rm]{subfig}
\usepackage{amssymb}
\usepackage{natbib}

\usepackage{graphicx}
\usepackage{amsmath}
\usepackage{amsthm}
\usepackage{booktabs}
\usepackage{algorithm}
\usepackage{algorithmic}
\usepackage[switch]{lineno}

\newtheorem{assumption}{Assumption}

\newtheorem{definition}{Definition}

\usepackage{enumitem}

\def\Vbar{{\perp\!\!\!\perp}}

\newtheorem{theorem}{Theorem}

\title{Long-Term Individual Causal Effect Estimation via Identifiable Latent Representation Learning}

\author{
Ruichu Cai$^{1,2}$
\and
Junjie Wan $^1$
\and
Weilin Chen$^1$
\and
Zeqin Yang$^{1,4}$
\and
Zijian Li$^3$
\and
\\
Peng Zhen$^4$
\And
Jiecheng Guo$^4$
\\
\affiliations
$^1$School of Computer Science, Guangdong University of Technology, Guangzhou, China\\
$^2$Peng Cheng Laboratory, Shenzhen, China\\
$^3$Mohamed bin Zayed University of Artificial Intelligence, Masdar City, Abu Dhabi\\
$^4$DiDi China Ride Hailing Business Group, Beijing, China\\
\emails
\{cairuichu, wj1205131700, chenweilin.chn, youngzeqin, leizigin\}@gmail.com,
\{zhenpeng, jasonguo\}@didiglobal.com
}

\begin{document}

\maketitle

\begin{abstract}
Estimating long-term causal effects by combining long-term observational and short-term experimental data is a crucial but challenging problem in many real-world scenarios. 
In existing methods, several ideal assumptions, e.g. latent unconfoundedness assumption or additive equi-confounding bias assumption, are proposed to address the latent confounder problem raised by the observational data. However, in real-world applications, these assumptions are typically violated which limits their practical effectiveness. 
In this paper, we tackle the problem of estimating the long-term individual causal effects without the aforementioned assumptions. 
Specifically, we propose to utilize the natural heterogeneity of data, such as data from multiple sources, to identify latent confounders, 
thereby significantly avoiding reliance on idealized assumptions.
Practically, we devise a latent representation learning-based estimator of long-term causal effects.
Theoretically, we establish the identifiability of latent confounders, 
with which we further achieve long-term effect identification.
Extensive experimental studies, conducted on multiple
synthetic and semi-synthetic datasets, demonstrate the effectiveness of our proposed method.
\end{abstract}

\section{Introduction}
Estimating long-term causal effects is of increasing importance in many domains, such as healthcare, public education, marketing, and public policy \citep{hohnhold2015focusing,chetty2011does,fleming1994surrogate,zheng2025long}. In such long-term scenarios, it is usually difficult to conduct randomized control experiments to estimate the causal effects. Hence, a lot of researchers resort to the more easily accessible long-term observations. However, methods based on observational data still suffer from the latent confounding bias problem. Therefore, combining observational data and experimental data has emerged as a promising solution for estimating long-term causal effects \citep{imbens2024long,ghassami2022combining,hu2022identification}.
\begin{figure}[!ht] 
	\centering
	\subfloat[]{
		\includegraphics[width=0.15\textwidth]{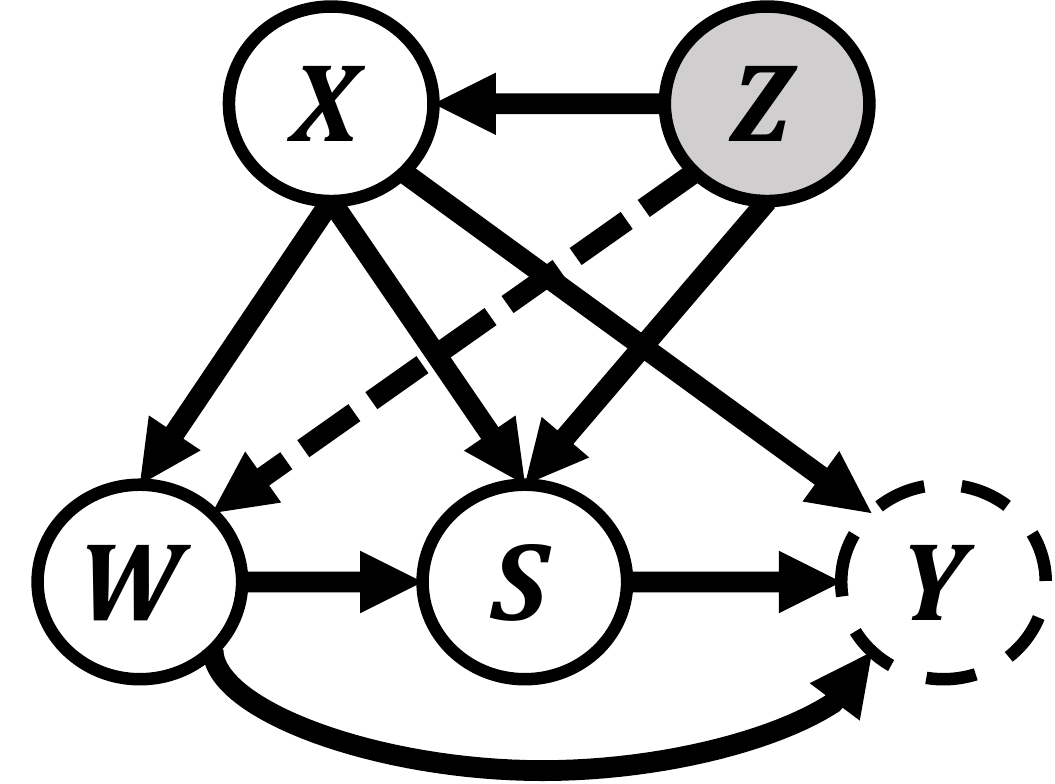} 
		\label{figure lat uncon}
	}
	\subfloat[]{
		\includegraphics[width=0.15\textwidth]{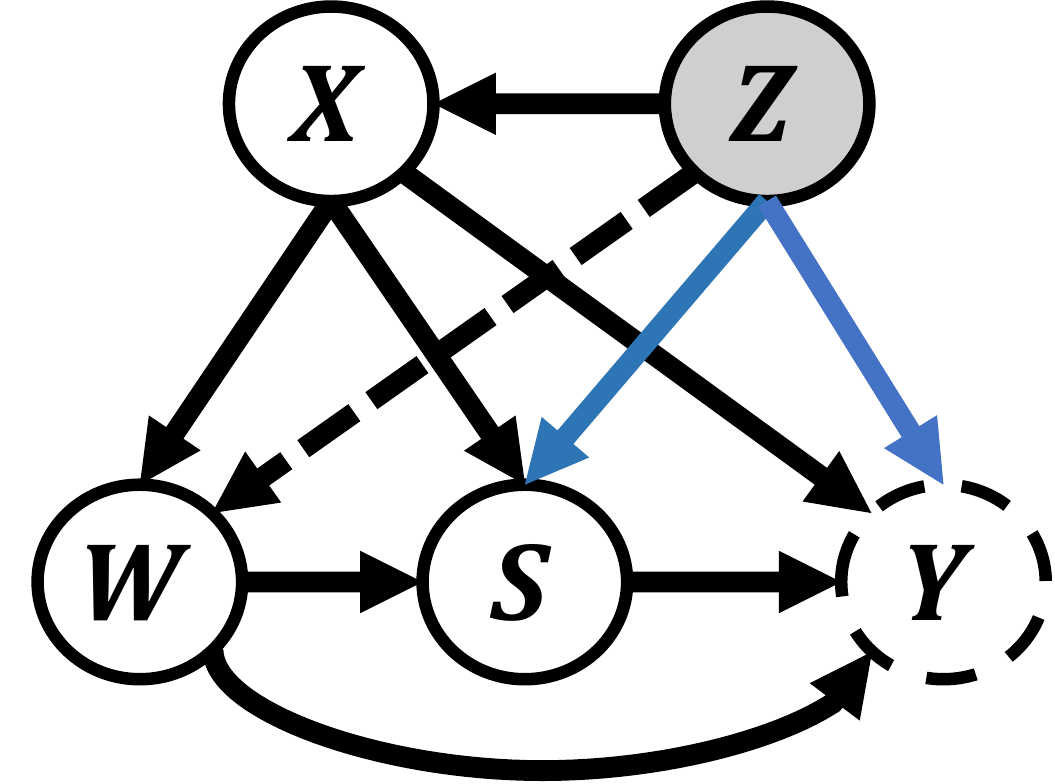}
		\label{figure condi uncon}
	}
   \subfloat[]{
		\includegraphics[width=0.15\textwidth]{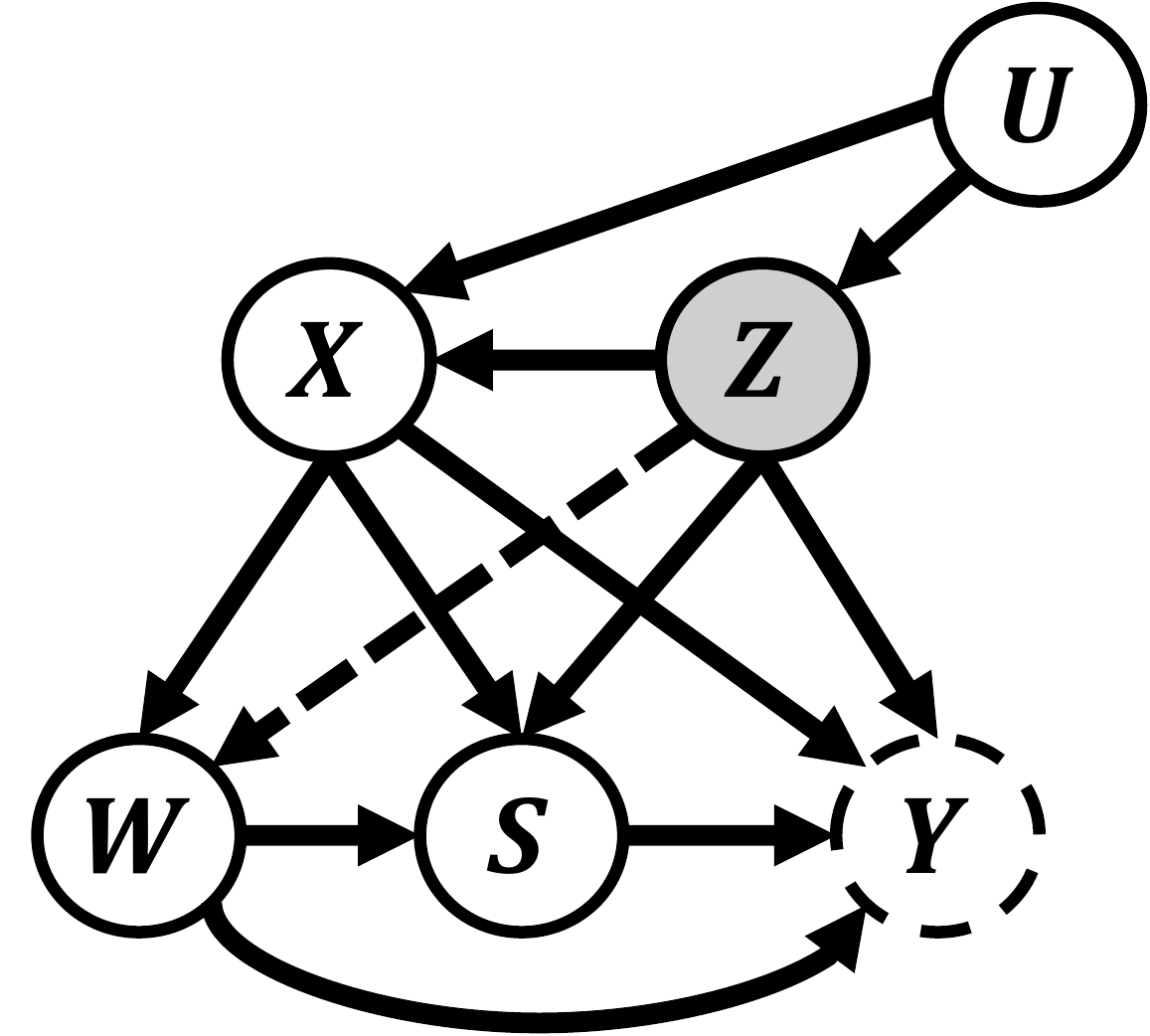}
		\label{figure our model setting graph}
	}
	\caption{Three causal graphs in long-term scenarios with $X$ being the pre-treatment variables, $Y$ being  the long-term outcome, $Z$ being the latent confounders, $S$ being short-term outcome, $U$ being the auxiliary variable, and $W$ being the treatment. White nodes denote the observed variables and grey nodes denote the unobserved variables. The dashed edges exist in the observational data but are absent in the experimental data. The dashed node $Y$ means $Y$ can be observed in observational data but not in experimental data. Specifically, Fig. \ref{figure lat uncon} shows the causal graph satisfying the latent confoundedness assumption \citep{athey2020combining}. Fig. \ref{figure condi uncon} shows the causal graph satisfying the equi-confounding bias assumption \citep{ghassami2022combining}, where the blue arrows in Fig. \ref{figure condi uncon} indicate the equal confounding bias. Fig. \ref{figure our model setting graph} shows the causal graph of our setting.
    }
     \label{fig:into causal graph} 
\end{figure}

Existing data combination-based methods estimate long-term effects mainly based on the so-called \textit{surrogate}.
As shown in Fig. \ref{fig:into causal graph}, the surrogate $S$ is the short-term outcome, serving as the supplement or replacement for the long-term outcome $Y$ in observational data.
However, the unconfoundedness assumption is usually violated in such observational data due to the existence of latent confounders $Z$. 
As a replacement for unconfoundedness assumption, \citeauthor{athey2020combining} propose an assumption named latent unconfoundedness, i.e., $Y(w)\perp W|X,S(w)$ on observational data, implicitly indicating the latent confounders $Z$ cannot affect long-term outcome $Y$ as illustrated in Fig. \ref{figure lat uncon}.
Alternatively, to relax the unconfoundedness assumption, \citeauthor{ghassami2022combining} introduces the (conditional) additive equi-confounding bias assumption, i.e.,  the magnitude of the confounding bias for the short-term and the long-term potential outcome variables are the same, as illustrated in Fig. \ref{figure condi uncon}.

Existing methods, however, encounter a \textbf{key challenge}: the ideal assumptions are usually violated in real-world applications, including both the latent unconfoundedness and additive equi-confounding bias assumptions, which limit their practical effectiveness. 
For example, in studying the effect of driver income (treatment $W$) on long-term retention (outcome $Y$) in a ride-hailing platform, driver characteristics (pre-treatment variable $X$) act as observed confounders affecting both income and retention.
However, the drivers' household expenses (latent confounders $Z$) may also affect drivers' long-term retention $Y$, violating the latent unconfoundedness assumption. 
Similarly, the additive equi-confounding bias assumption may be violated 
since household expenses can influence short- and long-term retention differently, i.e., the confounding bias varies over time rather than remaining constant. 
Therefore, the strong assumptions in existing methods still significantly limit their applicability.

To address the above challenge, we aim to develop a method without the above assumptions to estimate the individual long-term causal effects as shown in Fig. \ref{fig:causal graph}. 
Specifically, instead of assuming latent unconfoundedness or equi-confounding bias, we explore the identifiability of latent confounders $Z$ to estimate long-term causal effects.
To identify latent confounder $Z$, we resort to an additional auxiliary variable $U$, which is easily accessible from our readily available prior knowledge, such as the natural heterogeneity of data in real-world applications. 
Recall the aforementioned drivers' income study example, the data are usually collected from various cities, and the indicator variable of the city can be directly taken as the auxiliary variable. 
Leveraging the identifiability of $Z$, we establish the causal effect identification result and propose the corresponding latent representation learning-based estimator for long-term individual causal effects.
Overall, our contributions can be summarized as follows:
\begin{itemize}[leftmargin=8px]
    \item We focus on a more general setting for estimating long-term causal effects, as shown in Fig. \ref{figure our model setting graph}. 
    As shown in Fig. \ref{figure our model setting graph}, the assumed causal graph in our paper is a complete graph, and the causal graphs in existing work \citep{athey2020combining,ghassami2022combining} can be seen as our special cases.
    \item We theoretically achieve the identifiability of latent confounders. Leveraging the identifiability result, we further establish the identification of long-term individual effects.
    \item We devise a latent representation learning-based estimator for effect estimation. The effectiveness of our estimator is verified on five synthetic and two real-world datasets.
    \end{itemize}

\section{Related Works}

\textbf{Variational Auto-encoders for Causal Inference } 
Variational Auto-encoder (VAE) \citep{DBLP:journals/corr/KingmaW13} is a powerful tool to capture latent structure in different kinds of applications, e.g., image processing \citep{gregor2015draw} and time-series \citep{chung2015recurrent, 10.1145/3696410.3714931}.
In causal inference, VAE is used to recover unobserved variables to achieve the identification and estimation of the effects. Without unconfoundedness assumption,
CEVAE \citep{louizos2017causal} assumes that latent confounders can be recovered by their proxies and applies VAE to learn confounders.
As a follow-up work, TEDVAE \citep{zhang2021treatment} and DMAVAE \citep{xu2023disentangled} 
decouples the learned latent confounders into several factors to achieve a more accurate estimation of treatment effects in different settings.
With the recent development of VAE, nonlinear independent component analysis theory \citep{hyvarinen2016unsupervised} enables the identifiability of recovered variables, e.g., iVAE \cite{khemakhem2020variational} and SIG \citep{li2023subspace}.
CFDiVAE \citep{DBLP:conf/iclr/XuCLL0Y24} apply iVAE to recover the front-door adjustment variable, achieving effect identification under the front-door criterion \citep{pearl2009causality}. $\beta$-Intact-VAE \citep{wubeta} utilizes iVAE to recover prognostic scores to estimate effects under a limited overlap setting.
\textbf{Different} from them, we achieve long-term individual effect identification and estimation by applying iVAE to recover the latent confounders.

\textbf{Long-term Causal Inference} 
For decades, many works have explored what a valid surrogate is that can reliably predict long-term causal effects. 
Different types of criteria are proposed, e.g., prentice criteria \citep{prentice1989surrogate} and so on \citep{frangakis2002principal,lauritzen2004discussion}.
Recently, many works have explored estimating long-term causal effects based on surrogates via data combination. 
Under the unconfoundedness assumption, LTEE \citep{cheng2021long} and Laser \citep{cai2024long} are based on different designed neural networks for long-term causal inference.
EETE \citep{kallus2024role} studies the data efficiency from the surrogate and proposes efficient treatment effect estimation.
Some works \citep{pmlr-v235-wu24x,yang2024learning} also focus on balancing short- and long-term rewards under the unconfoundedness assumption.
Under surrogacy assumption, SInd \citep{athey2019surrogate} constructs the Surrogate Index as the substitutions for long-term outcomes in the experimental data to achieve effect identification and \citep{singh2022generalized} propose a kernel ridge regression-based estimator for long-term effect under continuous treatment.
As follow-up work, \citep{athey2020combining} assumes latent unconfoundedness assumption, i.e., short-term potential outcomes can mediate the long-term potential outcomes, to identify long-term causal effects. Under this assumption, several methods \citep{yang2024estimating, chen2023semiparametric} are proposed to more accurately estimate long-term effects.
Other feasible assumptions are proposed to replace the latent unconfoundedness assumption, e.g., the additive equi-confounding bias assumption \citep{ghassami2022combining,chen2025nonparametric} and its variant \citep{chen2025long}.
Based on proximal methods, the sequential structure surrogates are studied \citep{imbens2024long}.
\textbf{Different} from them, we focus on estimating long-term individual causal effects in a more general scenario as shown in Fig. \ref{figure our model setting graph}.

\section{Problem Definition}

\subsection{Notations}
Our notations follow the potential outcome framework \citep{rubin1974estimating}.
Let $W\in \{0,1\}$ be a binary treatment variable.
Let $d_\circ$ be the dimension of variable $\circ$.
Let $X\in \mathcal X \subseteq \mathbb{R}^{d_x}$ be pre-treatment variable, 
$Z\in \mathcal Z \subseteq \mathbb{R}^{d_z}$ be latent confounders, 
$S\in \mathcal S \subseteq  \mathbb{R}^{d_s}$ be the short-term outcome variable, $Y\in \mathcal Y \subseteq  \mathbb{R}$ be the long-term outcome variable, 
and $U\in  \mathcal U \subseteq \mathbb{R}^{d_u}$ be the auxiliary variable.
Further, we denote the potential short-term outcomes $S(w)\in \mathbb{R}^{d_s}$ and potential long-term outcomes $Y(w)\in \mathbb{R}$.
Denote $G \in \{o, e\}$ be the indicator of the data group, where $G = o$ indicates the observational data, and $G = e$ indicates the experimental data.  
Let lowercase letters (e.g., $x,y$) denote the value of random variables. 
Let lowercase letters with superscript $(i)$ denote the value of the specified $i$-th unit.
Following existing work \citep{athey2020combining,hu2022identification,ghassami2022combining}, we consider the data combination setting.
We have two types of data: the experimental data 
$\mathbb{D}_{exp}=\{x^{(i)},w^{(i)},s^{(i)},u^{(i)},g^{(i)}=e\}^{n_e}_{i=1}$
and the observational data 
$\mathbb{D}_{obs}=\{x^{(i},w^{(i)},s^{(i)},y^{(i)},u^{(i)},g^{(i)}=o\}^{n_o}_{i=1+n_e}$,
where $n_e,n_o$ are the sample sizes of experimental and observational data respectively. 
Our setting is described in Fig. \ref{fig:causal graph}.

\begin{figure}[t] 
	\centering
	\subfloat[Observational data]{
		\includegraphics[width=0.15\textwidth]{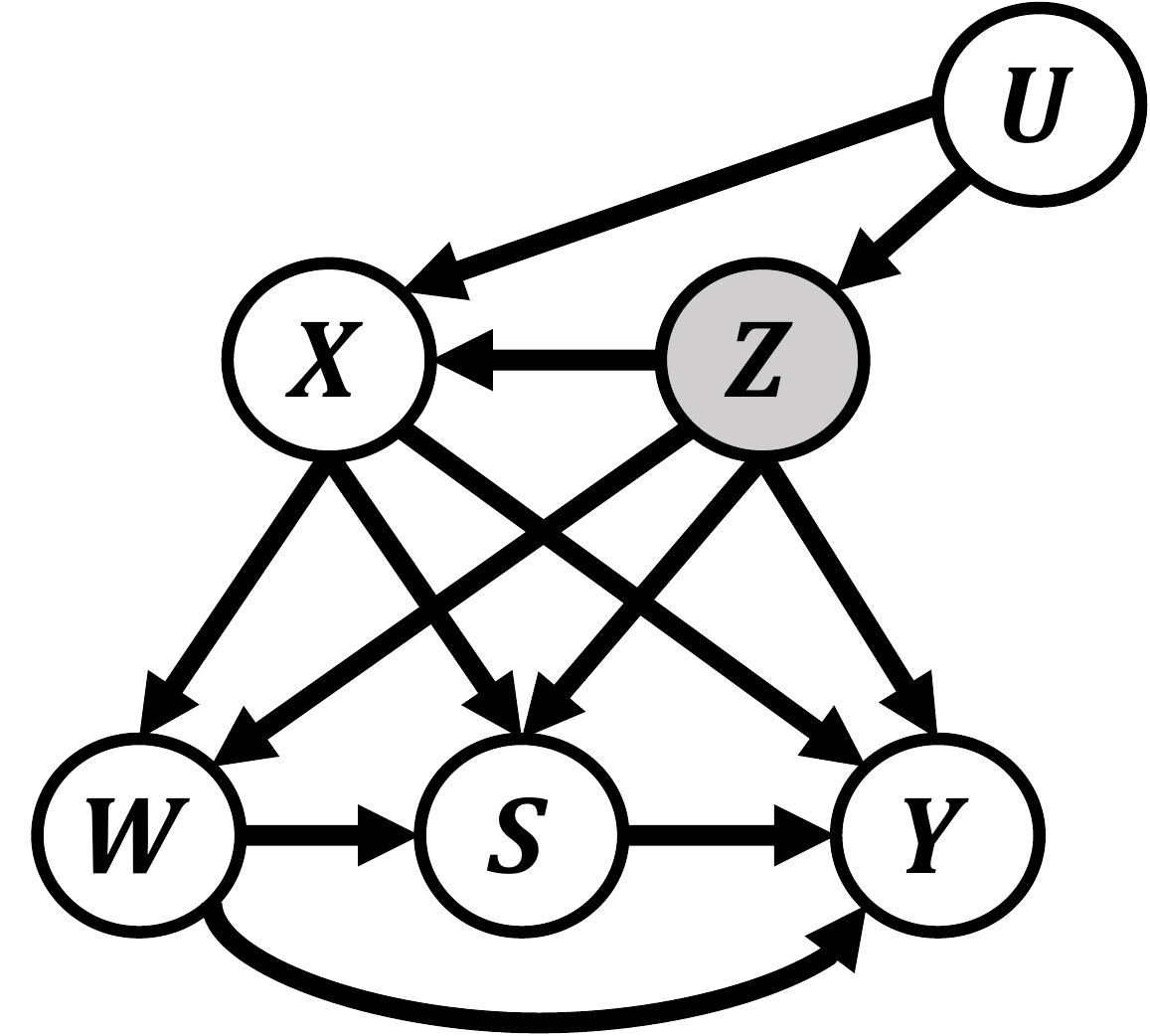} 
		\label{figure obs}
	}
	\hfil
	\subfloat[Experimental data]{
		\includegraphics[width=0.15\textwidth]{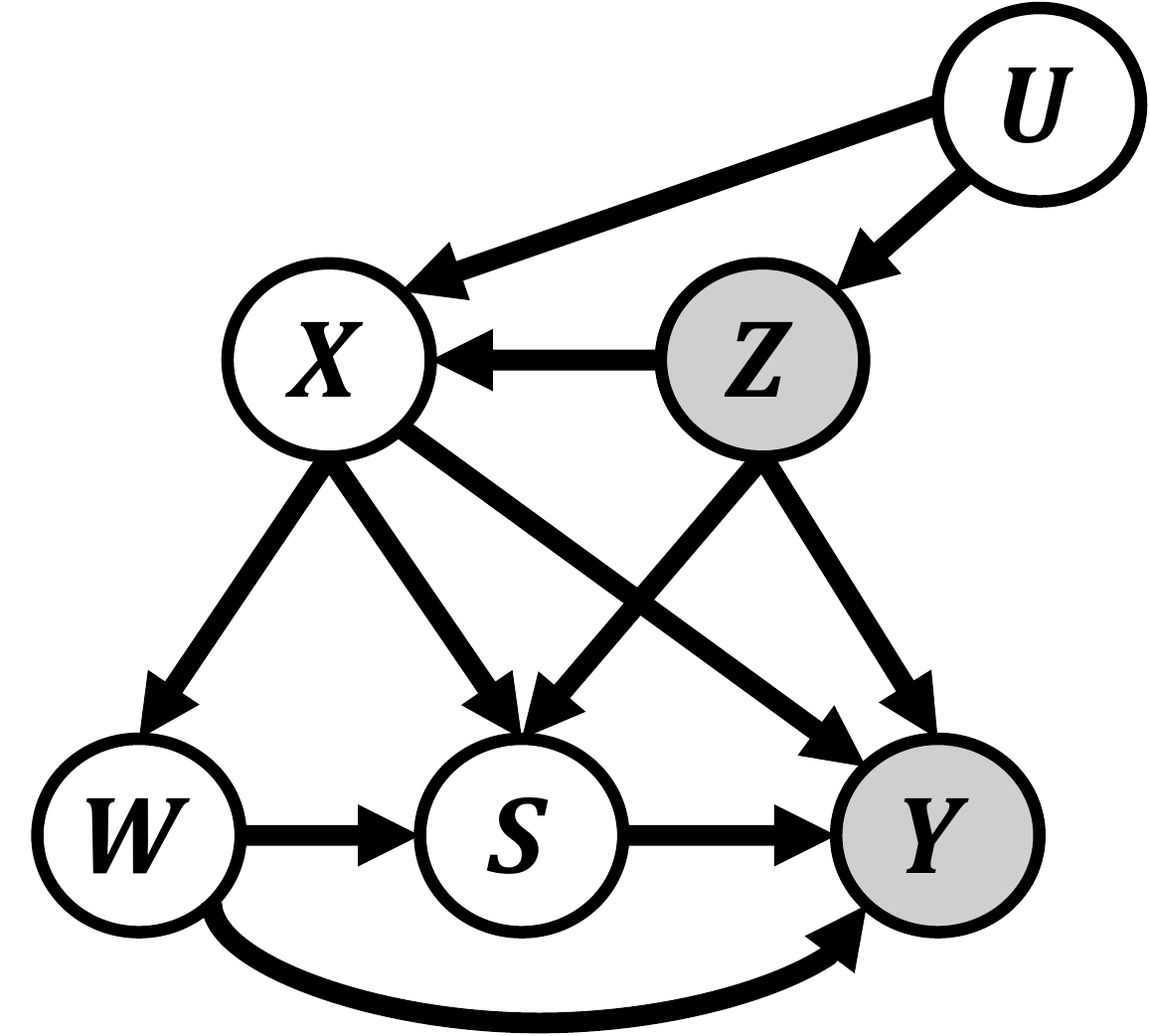}
		\label{figure exp}
	}
	\caption{Two causal graphs in our setting. The white nodes denote observed variables and the grey denote unobserved variables. Fig. \ref{figure obs} is the causal graphs of observational in our setting. Fig. \ref{figure exp} is the causal graph of experimental data in our setting.}
	 \label{fig:causal graph} 
\end{figure}

\subsection{Assumptions and Target Estimands}

Throughout this paper, we make the following assumptions:

\begin{assumption}[Long-term Effect Identification Assumptions] \citep{athey2020combining,ghassami2022combining} \label{assum: long term total}
\begin{enumerate}[label=\textbf{A\arabic*}, ref=A\arabic*]
    \item \textbf{[Consistency, Positivity]} If $W=w$, then $Y=Y(w)$ and $S=S(w)$. $\forall w,x$, $0<P(W=w|X=x)<1,~ 0<P(G=o|W=w, X=x)<1$.  \label{assum: consist, positi} 
    \item  \textbf{[Weak internal validity of observational data]} 
     for all $ w\in\{0,1\} $, $W\Vbar \{Y(w), S(w)\}|X, Z, G=o$. \label{assum: internal validity of obs} 
    \item  \textbf{[Internal validity of experimental data]}
    for all $ w\in\{0,1\} $, $W\Vbar \{Y(w), S(w)\}| X, G=e$. 
     \label{assum: internal validity of exp} 
    \item  \textbf{[External validity of experimental data]} 
    for all $ w\in\{0,1\} $, $G\Vbar \{Y(w), S(w)\}| X$. 
     \label{assum: external validity of exp} 
\end{enumerate}
\end{assumption}

The assumptions above are mild and widely used in existing literature, e.g.,  \cite{athey2020combining,ghassami2022combining}.
\ref{assum: consist, positi} is a standard assumption.
\ref{assum: internal validity of obs} allows the existence of latent confounders $Z$. \ref{assum: internal validity of exp} guarantees that the experimental data is unconfounded conditioned on $X$.
\ref{assum: external validity of exp} allows us to generalize the conditional distribution of potential outcomes between observational and experimental data.

In this paper, our \textbf{task} is to estimate the long-term individual treatment effects (ITE) given $\mathbb{D}_{exp},\mathbb{D}_{obs}$, defined as:
\begin{equation} 
  \tau(x)=\mathbb{E}[Y(1)-Y(0)|X=x],
\label{ITE_expression}
\end{equation}
as well as long-term average treatment effects (ATE), defined as:
\begin{equation} 
  \tau=\mathbb{E}[\tau(x)].
\label{ATE_expression}  
\end{equation}

\begin{figure*}[t] 
	\centering
	\subfloat[Inference network]{
		\includegraphics[width=0.4\textwidth]{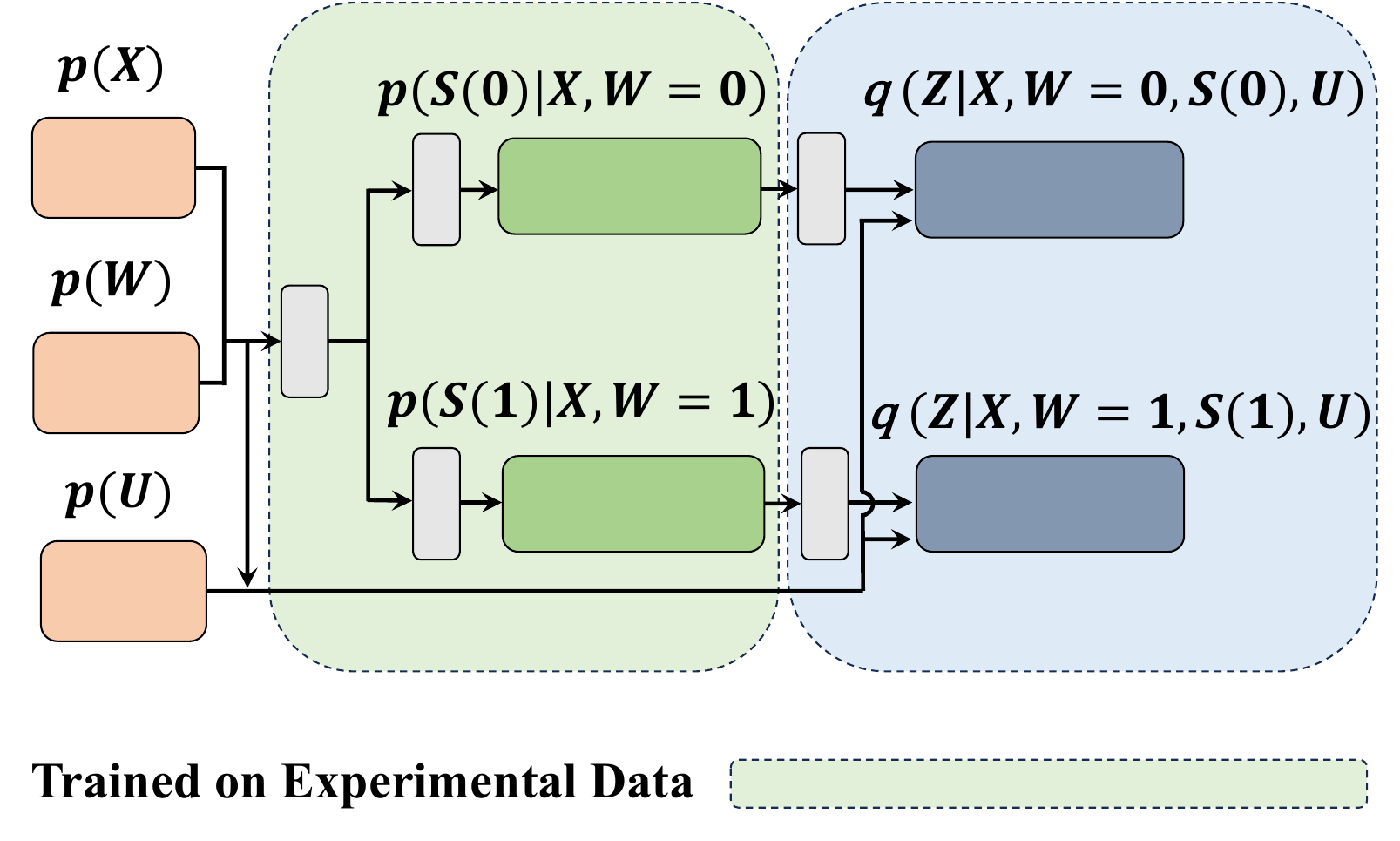} 
		\label{figure encoder}
	}
	\hfil
	\subfloat[Generative network]{
		\includegraphics[width=0.47\textwidth]{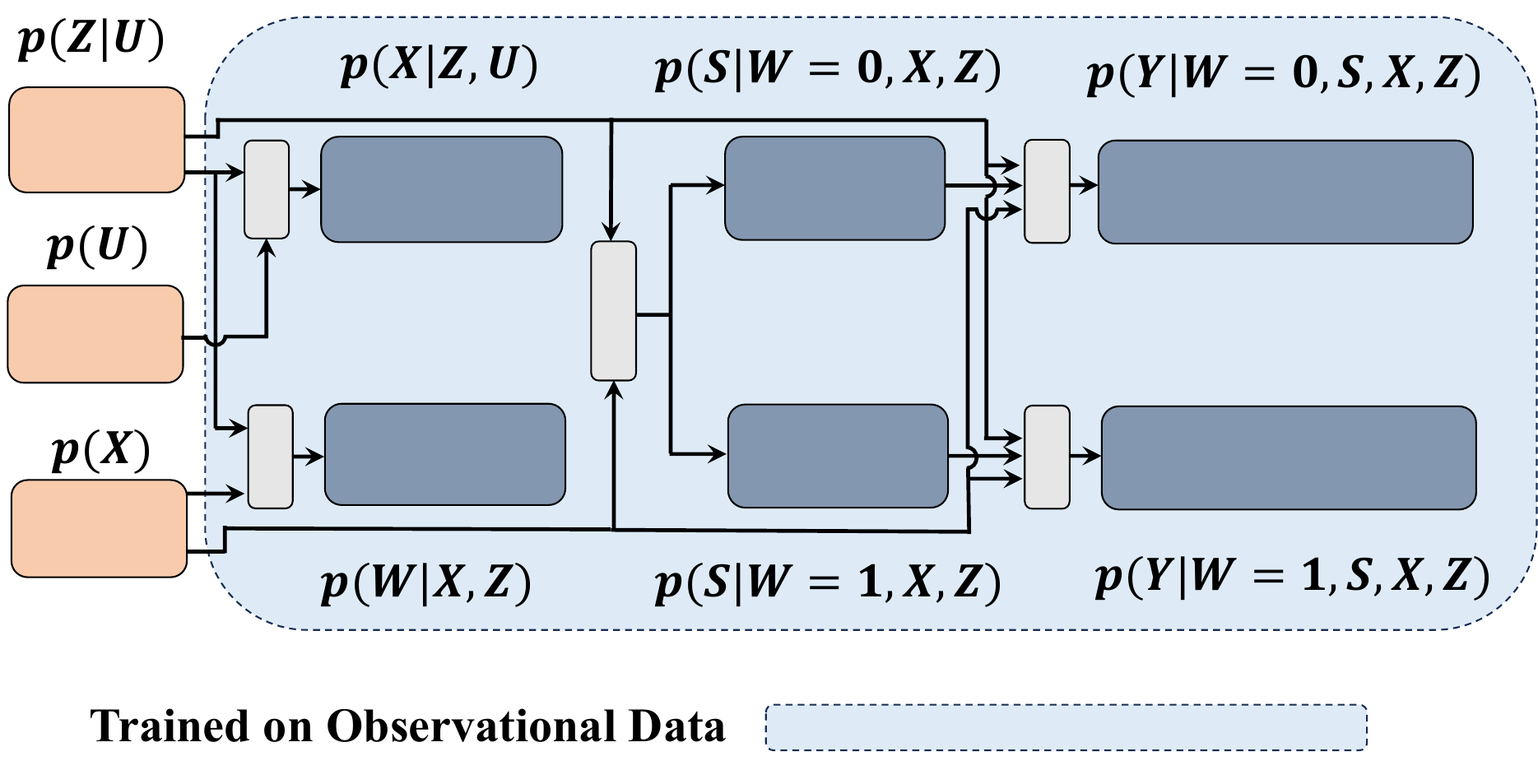}
		\label{figure decoder}
	}
	\caption{Overall architecture of the generative and inference networks for our model. Grey nodes represent MLP, green nodes correspond to the distribution trained on experimental data and blue nodes correspond to the distribution trained on observational data.}
	 \label{fig:model} 
\end{figure*}

\section{Methodology}

In this section, we present our end-to-end long-term causal effect estimator. Overall, as shown in Fig. \ref{fig:model}, 
our estimator consists of three modules: short-term potential outcome estimation,  latent representation learning, and ITE estimation. 
In the short-term potential outcome estimation module, we train an estimator for  $p(S(w)|W,X)$ using experimental data, as it is identifiable as $p(S|W,X)$.
In the latent representation learning module, we leverage variational inference to learn the latent representation of confounders $Z$. The pre-treatment variable $X$, treatment $W$ and the short-term potential outcome $S(w)$, obtained from the short-term potential outcome estimation module, are jointly treated as proxies for $Z$, ensuring sufficient information is available to recover $Z$. Additionally, the auxiliary variable $U$ is used as a prior, guaranteeing the identifiability of the latent confounder $Z$, as demonstrated in the theoretical analysis (see Section \ref{sec:Theoretical Guarantee}).

In the ITE estimation module, based on learned $Z$, we conduct an estimator to learn the potential outcomes in treated and control groups, whose difference results in the final estimator of $\tau(x)$.
Note that, the first module is trained on experimental data to ensure the identification of short-term potential outcomes, and the others are trained on observational data since the long-term outcome is only observed in observational data.

\subsection{Short-term Potential Outcome Estimation}

We employ a multilayer perceptron (MLP) to model the distribution of $p(S(w)|X)$ as our short-term potential outcome estimator.
Since we can access short-term experimental data, $p(S(w)|X)$ can be rewritten as $p(S|X,W=w)$ on experimental data. To estimate that, inspired by Tarnet \citep{johansson2022generalization}, we use two heads of MLP for the estimation. Specifically, we can model each dimension of $S(w)$ as a Gaussian distribution as follows:
\begin{equation}
    p(S|W,X)=\prod \limits_{i=0}^{d_s} \mathcal{N} (\mu=\hat{\mu}_{S_i}, \sigma^2=\hat{\sigma}_{S_i}^2) 
    \label{q_s_wx dis},
\end{equation}
where $\hat{\mu}_{S_i}$ and $\hat{\sigma}_{S_i}$ are the mean and variance of the Gaussian distribution parametrized by the MLPs. We use the negative log-likelihood of Eq. \eqref{q_s_wx dis} as the objective function $\mathcal{L}_{S(w)}$ for the short-term potential outcome estimator as follows:
\begin{equation} \label{loss fun for s(w)}  
\begin{aligned}
     \mathcal{L}_{S(w)}= & -\mathbb{E}_{q_{\mathbb D_{exp}}}[\log{p(S(w)|X)}] \\
    = & -\mathbb{E}_{q_{\mathbb D_{exp}}}[\log{p(S|X,W)}],
\end{aligned}
\end{equation}
where $q_{\mathbb D_{exp}}$ is the empirical data distribution given by $\mathbb D_{exp}$.

\subsection{Latent
Representation Learning} \label{Latent Confounders Recovering}

In the latent
representation learning step, we employ iVAE to recover latent confounders $Z$, as shown in Fig. \ref{fig:model}. 
This module consists of two networks: an inference network and a generative network.
Specifically, for the inference network, the auxiliary variable $U$ serves as additional information and thus our prior distribution is $p(Z|U)$. We further use the posterior distribution $q(Z|S(1),S(0),X,W,U)$ to approximate the prior, where the short-term potential outcomes are obtained by the short-term potential outcome estimator discussed in the previous section.
For the generative network, we reconstruct the treatment $W$, the short-term outcome $S$ and the pre-treatment covariate $X$.

Following exiting VAE-based works \citep{louizos2017causal}, we choose the prior $p(Z|U)$ as Gaussian distribution:
\begin{equation} 
    p(Z|U)=\prod \limits_{i=0}^{d_z} \mathcal{N} (Z_i|\hat \mu_i, \hat \sigma_i^2) 
    \label{p_z_u dis },
\end{equation}
where $\hat \mu_i$ and $\hat \sigma_i$ are the mean and variance of the Gaussian distribution parametrized by the MLPs.

To approximate the prior, we model the posterior distribution $q(Z|S(1),S(0),X,W,U)$ as Gaussian distribution:
\begin{equation}
\begin{aligned}
    &q_0(Z|S(0),X,U)=\prod \limits_{i=0}^{d_z} \mathcal{N} (\mu=\hat{\mu}_{Z_i|W=0}, \sigma^2=\hat{\sigma}_{Z_i|W=0}^2), \\
    &q_1(Z|S(1),X,U)=\prod \limits_{i=0}^{d_z} \mathcal{N} (\mu=\hat{\mu}_{Z_i|W=1}, \sigma^2=\hat{\sigma}_{Z_i|W=1}^2), \\
    &q(Z|W,S(1),S(0),X,U)\\
    = & W \cdot q_1(Z|S(1),X,U)+(1-W) \cdot q_0(Z|S(0),X,U),
    \label{q_z_sxz(w) dis }
\end{aligned}
\end{equation}
where $\hat{\mu}_{Z_i|W=0}$ and $\hat{\sigma}_{Z_i|W=0}$ are the mean and variance of the Gaussian distribution parametrized by MLPs whose inputs are $X,W,U$ and estimated $S(W)$, and similarly for $\hat{\mu}_{Z_i|W=1}$ and $\hat{\sigma}_{Z_i|W=1}$.

In the generative network, for a continuous variable, we parametrize the distribution as a Gaussian with its mean and variance both given by MLPs. For a binary variable, we use a Bernoulli distribution parametrized by an MLP similarly. Thus, we employ the following distributions for $p(X|Z,U)$:
\begin{equation} 
\begin{aligned}    
     p(X|Z,U)&=\prod \limits_{i=0}^{d_X} \mathcal{N} (\mu=\hat{\mu}_{X_i}, \sigma^2=\hat{\sigma}_{X_i}^2) \\
      \text{or} \quad 
     p(X|Z,U)&=\prod \limits_{i=0}^{d_X} \textbf{Bern}(\pi =\hat{\pi}_{X_i})
    \label{p_x_z dis },
\end{aligned}
\end{equation}
where $\hat{\mu}_{X_i}$ and $\hat{\sigma}_{X_i}$ are the mean and variance of the Gaussian distribution parametrized by MLPs in the generative network when the variable is continuous, and $\hat{\pi}_{X_i}$ is the mean of Bernoulli distribution parametrized by the generative network when the variable is binary. Similarly, we employ the following distributions for $p(W|X,Z)$ and $p(S|W,X,Z)$:
\begin{equation} 
\begin{aligned}   
    p(W|X,Z)&=\textbf{Bern}(\pi =\hat{\pi}_{W_i}),\\   
     p(S|W,X,Z)&=\prod \limits_{i=0}^{d_s} \mathcal{N} (\mu=\hat{\mu}_{S'_i}, \sigma^2=\hat{\sigma}_{S'_i}^2) \\
     \text{or} \quad 
     p(S|W,X,Z)&=\prod \limits_{i=0}^{d_s} \textbf{Bern}(\pi =\hat{\pi}_{S'_i})
    \label{p_w_xz dis&p_s_wxz dis },
\end{aligned}
\end{equation}
where $\hat{\mu}_{S'_i}$, $\hat{\sigma}_{S'_i}$, $\hat{\pi}_{S'_i}$ and $\hat{\pi}_{W_i}$ are all parametrized by the generative network. We then use the negative variational Evidence Lower Bound (ELBO) as the objective function for the inference and generative networks (see Appendix E for the derivations):
\begin{equation} 
\begin{aligned}    
     &\textbf{ELBO}
     =  \mathbb{E}_{q_{\mathbb D_{obs}} }[ \mathbb{E}_{q(Z|S(0),S(1),X,U,W)}[ \log{p(Z|U)} 
      \\ &
      + \log{p(X|Z,U)} +\log{p(W|X,Z)}
      +\log{p(S|W,X,Z)}
      \\
     &-\log{q(Z|S(0),S(1),X,U,W)}]],
    \label{elbo}
\end{aligned}
\end{equation}
where $q_{\mathbb D_{obs}}$ is the empirical data distributions given by $\mathbb D_{obs}$. 

\subsection{ITE Estimation} \label{subtitle ITE Estimation}

To obtain the outcome $Y$, we introduce an auxiliary distribution that helps predict long-term outcome $Y$. Specifically, we employ the following distribution for $p(Y|W,S,X,Z)$:
\begin{equation} 
\begin{aligned}   
     p(Y|W,S,X,Z)&= \mathcal{N} (\mu=\hat{\mu}_{y_i}, \sigma^2=\hat{\sigma}_{y_i}^2),
    \label{p_y_sxz dis }
\end{aligned}
\end{equation}
where $\hat{\mu}_{y_i}$ and $\hat{\sigma}_{y_i}$ are the mean and variance of the Gaussian distribution parametrized by MLPs. We then use the negative log-likelihood as its objective function:
\begin{equation} \label{loss fun for Y}  
\begin{aligned}
     \mathcal{L}_{Y}= -\mathbb{E}_{q_{\mathbb D_{obs}}}[\mathbb{E}_{q(Z|S(0),S(1),X,U,W)}[\log{p(Y|W,S,X,Z)}]].
\end{aligned}
\end{equation}

Overall, our final objective function $\mathcal{L}$ is
\begin{equation} 
\begin{split}
     \mathcal{L}= &-\textbf{ELBO} + \mathcal{L}_{S(w)} +\mathcal{L}_Y.
    \label{loss fun}
\end{split}
\end{equation}

As a result, after training our method on experimental and observational data, given specific unit $x^{(i)}, u^{(i)}$, our final estimator yields long-term potential outcomes $\hat y(1)^{(i)}, \hat y(0)^{(i)}$ on the treated and control group respectively. Thus the estimated long-term individual effect of $x^{(i)},u^{(i)}$ is
\begin{equation} 
    \begin{aligned}
        \hat{\tau}(x^{(i)}) = \hat y(1)^{(i)}- \hat y(0)^{(i)}.
    \end{aligned}
\end{equation}

\section{Theoretical Analysis} \label{sec:Theoretical Guarantee}

In this section, we present the identifiability result of our model and the identification of long-term individual causal effects. 
If we can correctly identify the latent confounders $Z$, the long-term individual causal effect can be identified based on the learned representation of $Z$.
We first prove that $Z$ is identifiable up to a simple transformation. 
Leveraging the identifiability result of $Z$, we further prove that the long-term individual causal effect is identifiable.

\subsection{Identifiability of Latent Confounders}
\label{subtitle Identifiability Of Latent Confounders}

To clearly introduce the latent confounders identifiability result, we first denote $Z_i$ as the $i$-th dimension of $Z$.
The identifiability of latent confounders means that, for each ground-truth latent confounder $Z_i$, there exist a corresponding estimated latent confounder $\hat{Z_i}$ and an invertible function $h_i: \mathbb{R} \rightarrow \mathbb{R}$,
such that $Z_i = h_i(\hat{Z_j})$. Please refer to Appendix A for the formal definition of identifiability.

We show that latent confounders can be identified up to permutation and invertible component-wise transformations.
\begin{theorem}\label{th: z-ID}
Suppose the data-generation process follows Fig. \ref{fig:causal graph} and the following conditions hold:
\begin{itemize}[leftmargin=8px]
    \item Smooth and Positive Density: The probability density function of latent confounders is smooth and positive, i.e., $p_{Z|U}$ is smooth and $p_{Z|U} > 0$ over $\mathcal Z$ and $\mathcal U$.
    \item Conditional Independence: Conditioned on $U$, each $Z_i$ is independent, i.e., $\forall i, j \in \{1,...,d_z\}$, $i\not= j$, $\log{p_{Z|U}(Z|U)}=\sum^{d_z}_{i} q_i(Z_i,U) $ where $q_i$ is the log density of the conditional distribution, i.e., $q_i:=\log{p_{Z_i|U}}$.
    \item Linear Independence: For any $Z \in \mathcal Z \subseteq \mathbb{R}^{d_z}$, there exist $2d_z + 1$ values of $U$, i.e., $u_j$ with $j=0,1,...,2d_z$, such that the $2d_z$ vectors $w(Z, u_j) - w(Z, u_0)$ with $j = 1, ..., 2d_z$, are linearly independent, where vector $w(Z, U)$ is defined as follows:
    \begin{equation} 
        \begin{aligned}
        w(Z, U)= \Bigg(
        &\frac{\partial{q_{1}(Z_{1},U)}}{\partial{Z_{1}}}, \ldots, \frac{\partial{q_{d_z}(Z_{d_z},U)}}{\partial{Z_{d_z}}}, \\
        &\frac{\partial^{2}{q_{1}(Z_{1},U)}}{\partial{Z_{1}}^2}, \ldots, \frac{\partial^{2}{q_{d_z}(Z_{d_z},U)}}{\partial{Z_{d_z}}^2} 
        \Bigg).
        \end{aligned}
    \end{equation}
\end{itemize}
By modeling the aforementioned data generation process in Fig. \ref{fig:causal graph}, latent confounders $Z$ are identifiable.
\end{theorem}
The proof is given in Appendix C. 
The first two conditions are standard in the identifiability of existing nonlinear ICA works, e.g., \cite{kong2022partial,khemakhem2020variational}. 
More importantly, the third condition means that the auxiliary variable contains enough information, i.e., at least $2d_z+1$ distinct values of $U$. 
This assumption is plausible due to the nature of the heterogeneity of data, e.g.,  data from $11$ cities can ensure the identifiability of $Z$ with up to $5$ dimensions. Please refer to Appendix B for more implications of these conditions.

\subsection{Identifiability of Long-term ITE} \label{subtitle Identifiability Of ITE}

Building on the identifiability of latent confounders, in this section, we can further achieve the identification of long-term ITE.
As stated in Theorem \ref{th: z-ID}, the latent confounder $Z$ is identified up to simple invertible transformation, i.e., $\hat Z=h^{-1}(Z)$. 
Note the identifiability provides a fine-grained theoretical guarantee, ensuring all information of $Z$ is preserved. Thus, with the learned $\hat Z$, the long-term causal effects can be identified, as stated in the following theorem.

\begin{theorem}\label{th: ITE-ID}
Under Assumption \ref{assum: long term total}, 
suppose Theorem \ref{th: z-ID} hold, and then 
$\tau(x) = \mathbb E [Y(1)-Y(0)|X=x]$ is identifiable. 
\end{theorem}

The proof is given in Appendix D. 
Theorem \ref{th: ITE-ID} theoretically guarantees the correctness of our model, providing a feasible technology of long-term individual causal effects estimation via learning latent confounders.

\begin{table*}[!t] \small
    \centering
\renewcommand{\arraystretch}{1.2}
    \begin{tabular}{l|c c|c c|c c c@{}}
            \hline
         & \multicolumn{2}{c|}{Synthetic 1} & \multicolumn{2}{c|}{Synthetic 2} & \multicolumn{2}{c}{Synthetic 3} & \\ \cline{2-8}
         & $\epsilon_{ATE}$ & $\epsilon_{ITE}$ & $\epsilon_{ATE}$ & $\epsilon_{ITE}$ & $\epsilon_{ATE}$ & $\epsilon_{ITE}$ & \\ \hline
        CEVAE \citep{louizos2017causal} &$3.902_{\pm 0.740 }$ &$4.162_{\pm 0.781 }$ &$0.146_{\pm 0.037}$ &$0.270_{\pm 0.056}$ &$0.877_{\pm 0.161}$ &$0.975_{\pm 0.181}$\\ \hline
        TEDVAE \citep{zhang2021treatment} &$4.356_{\pm 1.078}$ &$4.851_{\pm 1.183}$ &$0.260_{\pm 0.109}$ &$0.397_{\pm 0.111}$ &$0.941_{\pm 0.186}$ &$ 1.171_{\pm 0.199}$\\ \hline
        LTEE \citep{cheng2021long} &$4.815_{\pm 1.269}$ &$5.726_{\pm 1.662}$ &$0.373_{\pm 0.232}$ &$0.596_{\pm 0.288}$ &$ 0.985_{\pm 0.174}$ &$1.215_{\pm 0.176}$ \\ \hline
        S-Learner \citep{kunzel2019metalearners} &$2.916_{\pm 0.854}$ &$4.185_{\pm 1.027}$ &$0.106_{\pm 0.171}$ &$0.500_{\pm 0.300}$ &$ 0.208_{\pm 0.159}$ &$2.235_{\pm 1.493}$ \\ \hline
        T-Learner \citep{kunzel2019metalearners} &$5.554_{\pm 2.733}$ &$7.687_{\pm 3.529}$ &$0.310_{\pm 0.308}$ &$0.746_{\pm 0.435}$ &$0.836_{\pm 0.917}$ &$1.832_{\pm 0.763}$ \\ \hline
        Imputaion \citep{athey2020combining}   &$2.480_{\pm 2.290}$ & - &$0.628_{\pm 0.542}$& - &$0.956_{\pm 1.094}$& - \\ \hline
        Weighting \citep{athey2020combining}  &$11.579_{\pm 6.775}$ & -&$1.896_{\pm 1.801}$ & - &$0.854_{\pm 0.901}$  & -\\ \hline
        Equi-naive \citep{ghassami2022combining} &$2.837_{\pm 1.377}$ &$ 4.297_{\pm 2.080}$ &$0.153_{\pm 0.145}$ &$ 0.974_{\pm 0.268}$&$\textbf{0.185}_{\pm 0.190}$ &$1.927_{\pm 0.443}$ \\ \hline
        IF-base \citep{ghassami2022combining} &$9.385_{\pm 7.690}$ & - &$1.600_{\pm 2.030}$ & - &$4.846_{\pm 3.716}$& - \\ \hline
        ICEVAE &$\textbf{2.402}_{\pm 0.436}$ &$\textbf{3.173}_{\pm 0.418}$ &$\textbf{0.105}_{\pm 0.068}$ &$\textbf{0.137}_{\pm 0.064}$  &$0.427_{\pm 0.385}$ &$\textbf{0.695}_{\pm 0.364}$ \\ \hline
    \end{tabular}
    \caption{
    Results of estimation error regarding ATE and ITE on three synthetic datasets. We report mean$\pm$std results. - means the method is not applicable. The best is bolded.
    }
    \label{Syn_result_table}
\end{table*}

\section{Experiments}

In this section, we verify the effectiveness of our model and the correctness of our theory. Specifically, we answer the following questions:
\begin{itemize}[leftmargin= 8px]
    \item[] 1. \textbf{Can our model identify latent confounders $Z$?}
    \item[] 2. \textbf{Does our model perform well on datasets that follow different existing assumptions?}
    \item[] 3. \textbf{Does our model outperform baselines on the real-world datasets?}
    \item[] 4. \textbf{Is our method robust to different strengths of latent confounding?}
\end{itemize}

\subsection{Experimental Setup}

\textbf{Datasets}
Since the ground-truth potential outcome can not be observed in the real world, following existing literature \citep{louizos2017causal,cheng2021long,cai2024long,yang2024estimating}, we use synthetic and semi-synthetic data to evaluate our method and baselines.

For the synthetic data, we simulate five synthetic datasets in our paper.
To validate the generalizability of our method, we first simulate three datasets corresponding to the causal graphs in Table \ref{Syn_result_table}. 
The first synthetic dataset allows all the existence of edges following the assumed causal graph in our paper. The second synthetic dataset follows the latent unconfoundedness assumption \citep{athey2020combining} that rules out the edges from unobserved confounders $Z$ to long-term outcome $Y$. The third dataset follows the additive equi-confounding bias assumption \citep{ghassami2022combining} that assumes the short-term confounding bias is equal to the long-term one. 
To further analyze the performances in terms of different strengths of confounding bias, we simulate the fourth synthetic dataset with varying $\beta$, which controls the coefficients in the data generation function from $Z$ to $W$ and $Z$ to $Y$. Finally, we simulate the fifth synthetic dataset to verify that our method is able to identify $Z$. 
All data generation details can be found in Appendix F.

For the semi-synthetic data, we use IHDP \citep{hill2011bayesian} and TWINS \citep{almond2005costs} to validate our model's performance on complex real-world data.
In detail, we reuse their original features and divide them into pre-treatment variables $X$, unobserved confounders $Z$ and the auxiliary variables $U$ according to their real-world meanings. Then we divide the samples into experimental and observational data and generate corresponding treatments, short-term outcomes, and long-term outcomes. The feature division and data generation details can be found in Appendix F.

\textbf{Baselines and Metrics} We compare our model \textbf{ICEVAE}\footnote{Code is available at https://github.com/DMIRLAB/ICEVAE and https://github.com/learnwjj/ICEVAE}  with the following baselines designed for long-term causal effect, including the \textbf{Imputation} and the \textbf{Weighting} approaches \citep{athey2020combining}, the naive estimator and the efficient influence function-based estimator under Conditional Additive Equi-Confounding Bias assumption \citep{ghassami2022combining}, named \textbf{Equi-naive} and \textbf{IF-based} respectively, and \textbf{LTEE} \citep{cheng2021long}. 
Besides, since there is a lack of work on estimating heterogeneous long-term causal effects, we use \textbf{CEVAE} \citep{louizos2017causal} as one of the baselines, as it is designed for recovered latent confounders in effects estimation.
We also compare our model with the follow-up work \textbf{TEDVAE} \citep{zhang2021treatment}. Finally, we introduce two simple estimators, the \textbf{S-Learner} and the \textbf{T-Learner} \citep{kunzel2019metalearners} to be baselines, which are implemented using MLPs. Note that the \textbf{Imputation} method, the \textbf{Weighting} method, and the \textbf{IF-based} method are designed for ATE and cannot estimate ITE.
The implementation details regarding baselines and our method can be found in Appendix F.

For metrics, 
to measure the error of average causal effect estimation, we report the mean and the standard deviation(std) of mean square error $\epsilon_{ATE}$ on the test set by performing 5 replications, i.e., $\epsilon_{ATE}=(\tau - \hat{\tau})^2$, where $\tau$ and $\hat{\tau}$ are the real and estimated average treatment effects on the test set respectively. 
To measure the error of estimating individual causal effects, we report the mean and std of Precision in the Estimation of Heterogeneous Effect (PEHE) $\epsilon_{ITE}$ on the test set by performing 5 replications 
where $ \epsilon_{ITE}
 = \frac{1}{n_{test}}\sum^{n_{test}}_{i=1}\left(\tau(x^{(i)})-\hat{\tau}(x^{(i)}) \right)^2$, where $n_{test}$ is the test sample size.

\addtocounter{table}{1}

\begin{table*}[!t]
\renewcommand{\arraystretch}{1.2}
 \resizebox{\textwidth}{!}{
    \begin{tabular}{@{}l|c c|c c|c c|c c|c c c@{}}
            \hline
         & \multicolumn{2}{c|}{$\beta=1$} & \multicolumn{2}{c|}{$\beta=1.5$} & 
         \multicolumn{2}{c|}{$\beta=3$} & 
         \multicolumn{2}{c|}{$\beta=4.5$} & \multicolumn{2}{c}{$\beta=5$} & \\ \cline{2-12}

         & $\epsilon_{ATE}$ & $\epsilon_{ITE}$ & $\epsilon_{ATE}$ & $\epsilon_{ITE}$& $\epsilon_{ATE}$ & $\epsilon_{ITE}$& $\epsilon_{ATE}$ & $\epsilon_{ITE}$& $\epsilon_{ATE}$ & $\epsilon_{ITE}$ & \\ \hline
        CEVAE 
        &$0.116_{\pm{0.083}}$
        &$0.188_{\pm{0.081}}$
        &$0.324_{\pm{0.115 }}$
        &$0.401_{\pm{0.119}}$
        &$3.902_{\pm{0.740}}$
        &$4.162_{\pm{0.781}}$
        &$14.348_{\pm{3.036}}$
        &$15.406_{\pm{3.493}}$
        &$19.403_{\pm{3.152}}$
        &$20.444_{\pm{3.585}}$
        \\ \hline
        TEDVAE 
        &$0.097_{\pm{0.039 }}$
        &$0.205_{\pm{0.041 }}$
        &$0.351_{\pm{0.065 }}$
        &$0.498_{\pm{0.068}}$
        &$4.356_{\pm{1.078}}$
        &$4.851_{\pm{1.183}}$
        &$16.543_{\pm{3.757 }}$
        &$18.589_{\pm{4.633}}$
        &$22.046_{\pm{3.856 }}$
        &$24.036_{\pm{4.571}}$
        \\ \hline
        LTEE 
        &$0.048_{\pm{0.034 }}$
        &$0.212_{\pm{0.071 }}$
        &$0.296_{\pm{0.141 }}$
        &$0.510_{\pm{0.220}}$
        &$4.815_{\pm{1.269}}$
        &$5.726_{\pm{1.662}}$
        &$17.678_{\pm{ 7.534}}$
        &$20.310_{\pm{9.404}}$
        &$23.980_{\pm{7.371 }}$
        &$26.740_{\pm{9.062 }}$
        \\ \hline
        S-Learner 
        &$\textbf{0.021}_{\pm{0.013}}$
        &$0.414_{\pm{0.091  }}$
        &$\textbf{0.096}_{\pm{0.090 }}$
        &$0.617_{\pm{0.098}}$
        &$2.916_{\pm{0.854}}$
        &$4.186_{\pm{1.027}}$
        &$15.382_{\pm{7.409 }}$
        &$19.080_{\pm{8.705}}$
        &$18.842_{\pm{9.137 }}$
        &$22.609_{\pm{ 10.251}}$
        \\ \hline
        T-Learner 
        &$0.190_{\pm{0.130 }}$
        &$0.582_{\pm{0.193 }}$
        &$0.209_{\pm{0.211 }}$
        &$0.867_{\pm{0.304}}$
        &$5.554_{\pm{2.733}}$
        &$7.687_{\pm{3.529}}$
        &$17.026_{\pm{6.587 }}$
        &$21.598_{\pm{7.321}}$
        &$20.068_{\pm{10.22}}$ 
        &$26.478_{\pm{9.314}}$
        \\ \hline
        Imputation
        &$0.792_{\pm{0.934 }}$
        &-
        &$0.928_{\pm{1.258 }}$
        &-
        &$2.480_{\pm{2.290}}$
        &-
        &$13.156_{\pm{7.144 }}$
        &-
        &$19.518_{\pm{ 11.240}}$
        &-
        \\ \hline
        Weighting 
        &$0.861_{\pm{0.648 }}$
        &-
        &$0.639_{\pm{0.339 }}$
        &-
        &$11.579_{\pm{6.775}}$
        &-
        &$51.634_{\pm{13.346 }}$
        &-
        &$70.104_{\pm{ 11.687}}$
        &-
        \\ \hline
        Equi-naive
        &$0.285_{\pm{0.422 }}$
        &$0.823_{\pm{0.353} }$
        &$0.247_{\pm{0.272} }$
        &$1.001_{\pm{0.245}}$
        &$2.837_{\pm{1.377}}$
        &$4.297_{\pm{2.080}}$
        &$10.619_{\pm{14.245 }}$
        &$14.245_{\pm{7.583}}$
        &$19.314_{\pm{3.186}}$
        &$22.978_{\pm{3.406}}$
        \\ \hline
        IF-base 
        &$0.619_{\pm{0.831 }}$
        &-
        &$1.707_{\pm{1.893 }}$
        &-
        &$9.385_{\pm{7.690}}$
        &-
        &$30.562_{\pm{13.998 }}$
        &-
        &$32.723_{\pm{12.970 }}$
        &-
        \\ \hline
        ICEVAE
        &$0.038_{\pm{0.031 }}$
        &$\textbf{0.069}_{\pm{0.032} }$
        &$0.182_{\pm{0.086}}$
        &$\textbf{0.217}_{\pm{0.085}}$
        &$\textbf{2.402}_{\pm{0.436}}$
        &$\textbf{3.173}_{\pm{0.418}}$
        &$\textbf{9.897}_{\pm{2.685}}$
        &$\textbf{11.395}_{\pm{2.841}}$
        &$\textbf{14.960}_{\pm{3.363}}$
        &$\textbf{16.467}_{\pm{3.935}}$
        \\ \hline
    \end{tabular}
    }
    \caption{Results of estimation error regarding ATE and ITE on the fourth synthetic dataset with different strengths of confounding bias controlled by $\beta$. We report mean$\pm$std results. - means the method is not applicable. The best is bolded.
    }
    \label{Dif_beta_Syn_result_table}
\end{table*}

\addtocounter{table}{-2}

\subsection{Results and Analysis}

\subsubsection{Can our model identify latent confounders $Z$?}
\begin{figure}[!h] 
	\centering
	\subfloat[Scatterplots of $Z_1$ and $\hat Z_1$.]{
		\includegraphics[width=0.2\textwidth]{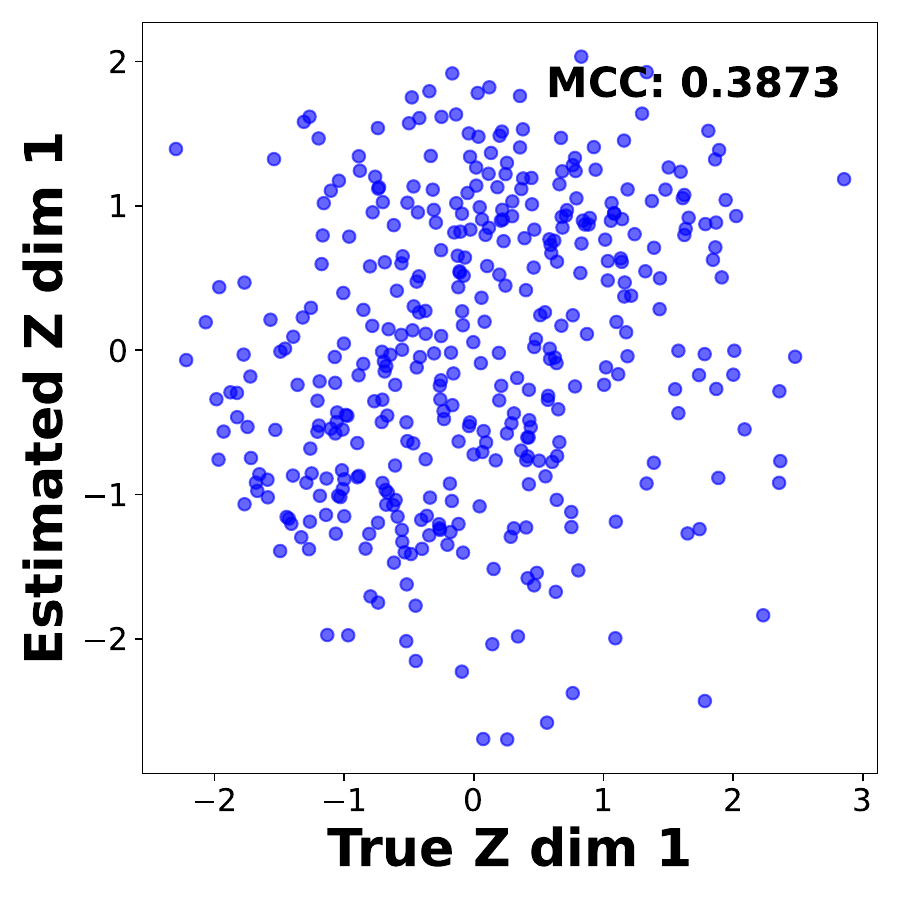} 
		\label{figure MCC1}
	}
	 \hfil \hfil \hfil \hfil 
	\subfloat[Scatterplots of $Z_1$ and $\hat Z_2$.]{
		\includegraphics[width=0.2\textwidth]{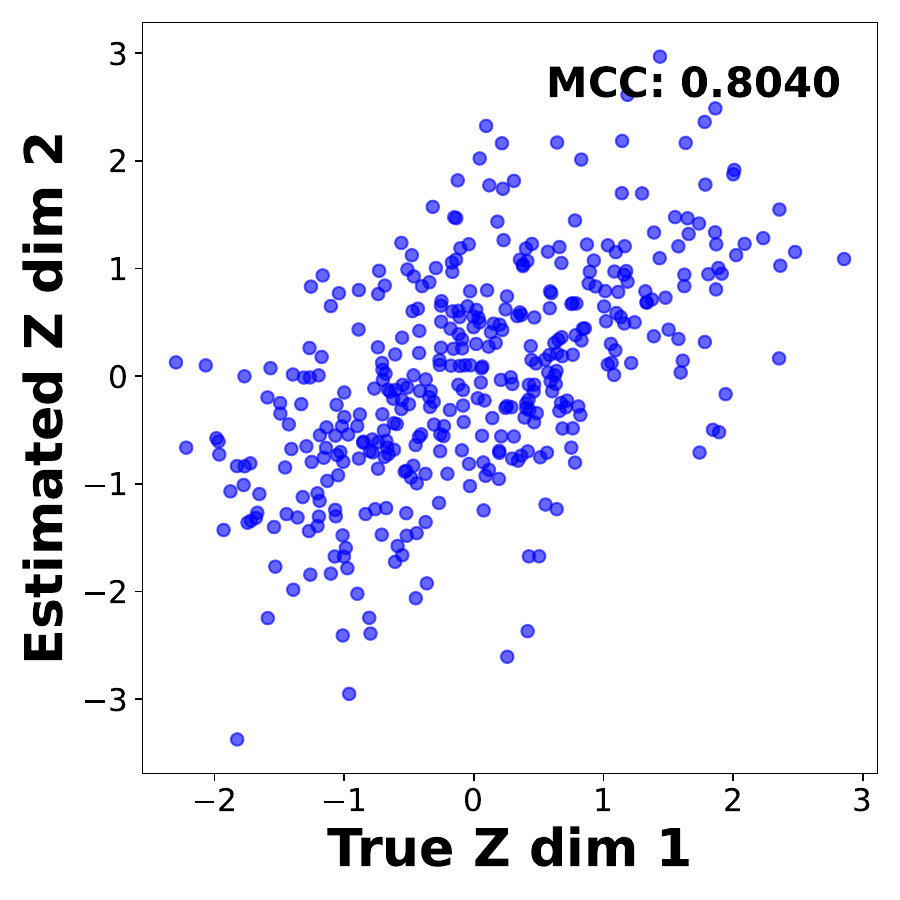}
		\label{figure MCC2}
	}
    \hfil
    \subfloat[Scatterplots of $Z_2$ and $\hat Z_1$.]{
		\includegraphics[width=0.2\textwidth]{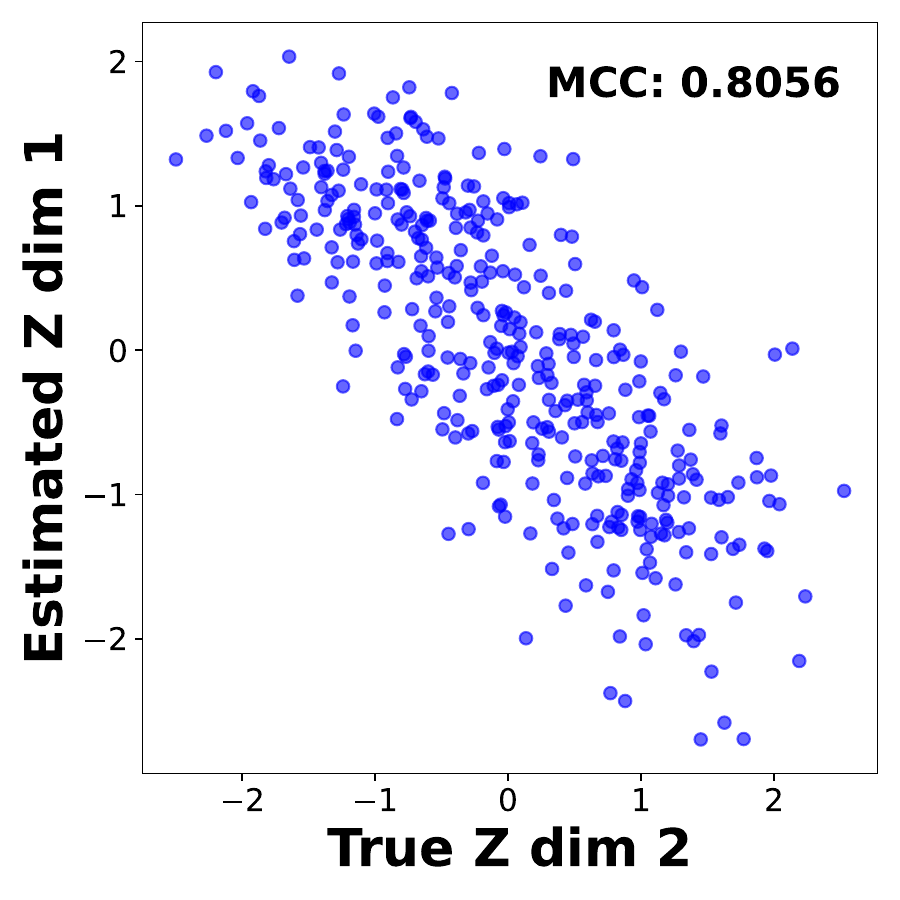} 
		\label{figure MCC3}
	}
	\hfil \hfil \hfil \hfil 
	\subfloat[Scatterplots of $Z_2$ and $\hat Z_2$.]{
		\includegraphics[width=0.2\textwidth]{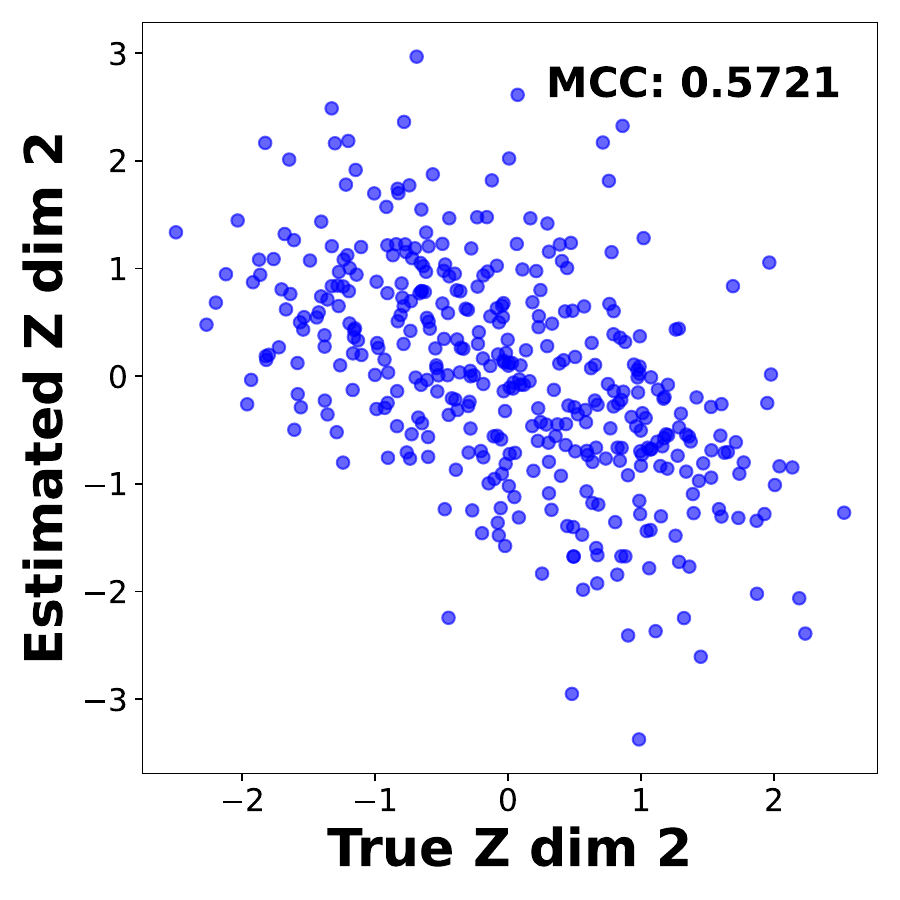}
		\label{figure MCC4}
	}
	\caption{Result on the fifth synthetic dataset. Fig. \ref{figure MCC1}-\ref{figure MCC4} show the scatterplots between each ground-truth and estimated latent confounder.}
	 \label{fig:recover_MCC} 
    \vspace{-.3cm}
\end{figure}

To validate the correctness of Theorem \ref{th: z-ID}, we conduct experiments by applying our method to the \textbf{Synthetic 5} dataset.
As shown in Fig. \ref{fig:recover_MCC},
the latent variables are successfully recovered, with a high MCC metric calculated by the ground-truth $Z$ and estimated $Z$. 
Fig. \ref{fig:recover_MCC} suggests that the latent causal variables are estimated up to permutation and component-wise invertible transformation, i.e., the estimated $Z_1$ in the figure corresponds to the true $Z_2$, with an MCC value of 0.8056. The estimated $Z_2$ corresponds to the true $Z_1$, with an MCC value of 0.8040. 
This indicates that our proposed method is able to identify $Z$, which verifies the correctness of our Theorem \ref{th: z-ID}.
\vspace{-.2cm}
\subsubsection{Does our model perform well on datasets that follow different existing assumptions?}
We conduct experiments by comparing our method with baselines on three different synthetic datasets that follow different data generation processes.
The results are shown in Table \ref{Syn_result_table}.
Overall, on all three datasets, our method achieves almost the best performance, revealing the generalizability of our method under different assumptions.
In detail, on the Synthetic 1 dataset, our method achieves the lowest ITE and ATE estimation error and std, indicating the effectiveness of our method.
As for the results of the Synthetic 2 dataset, compared with baselines, our method achieves comparable performance. Note that the Imputation and Weighting methods perform much better on the Synthetic 2 dataset than the Synthetic 1 dataset since the Synthetic 2 dataset is designed following the latent unconfoundedness assumption.
Similarly, as for the results of the Synthetic 3 dataset that is generated following the additive equi-confounding bias assumption, Equi-naive can achieve the lowest error in terms of ATE estimation. On this dataset, our method also achieves comparable performance, especially in terms of ITE estimation.
Hence, we conclude that our model can perform well on datasets that follow different existing assumptions.


\subsubsection{Does our model outperform baselines on real-world datasets?}
In Table \ref{Semi_result_table}, we evaluate the performance of our model on complex real-world data by comparing each method using two semi-synthetic datasets. 
The main observations are as follows. 
Overall, our method achieves the best performance regarding ITE estimation and comparable performance regarding ATE estimation, indicating the effectiveness of our method.
Specifically, compared with the VAE-based method, our method performs better, which indicates that the experimental data does help recover latent confounders.
Compared with the Imputation and Weighting methods, our method strongly outperforms them, since the unsuitable latent unconfoundedness assumption is made by their methods.
In conclusion, we find that our method ICEVAE can outperform baselines on real-world datasets.
\begin{table}[!t]
\centering \small
\renewcommand{\arraystretch}{1.2}
 \resizebox{.48\textwidth}{!}{
\begin{tabular}{@{}l|c c | c c @{}}
        \hline
         & \multicolumn{2}{c|}{IHDP}  & \multicolumn{2}{c}{TWINS}  \\ \hline
         & $\epsilon_{ATE}$ & $\epsilon_{ITE}$ &  $\epsilon_{ATE}$ & $\epsilon_{ITE}$  \\ \hline
        CEVAE &$\textbf{0.004}_{\pm0.003}$ & $0.183_{\pm 0.054}$ & $0.641_{\pm0.521}$ & $16.818_{ \pm 11.867}$  \\ \hline
        TEDVAE & $0.011_{\pm0.019}$ &$0.188_{\pm 0.032}$ & $ 0.824_{ \pm1.108 }$& $16.657 _{ \pm 11.978 }$ \\ \hline
        LTEE &$0.015_{\pm0.014}$ & $0.668_{\pm0.132 }$& $1.994_{ \pm2.470}$ & $15.667_{ \pm 14.297 }$\\ \hline
        T-Leaner &$0.061_{\pm0.051 }$& $1.060_{\pm0.214}$ & $4.665_{ \pm5.505 }$& $5.191_{ \pm5.581  }$ \\ \hline
        S-Leaner &$0.020_{\pm0.018} $&$0.969_{\pm0.354}$  & $2.536_{ \pm4.101 }$& $12.288_{ \pm4.779 }$ \\ \hline
        Imputaion  &$0.713_{\pm0.478}$ &-& $46.092_{ \pm43.729}$ &- \\ \hline
        Weighting &$0.664_{\pm0.959}$ &- & $6.597_{ \pm10.859}$ &- \\ \hline
        ICEVAE & $0.016_{\pm0.027}$ & $\textbf{0.178}_{\pm 0.060}$& $\textbf{0.204}_{ \pm0.229}$& $ \textbf{3.665} _{ \pm 2.246} $ \\ \hline
    \end{tabular}
    }
    \caption{Results of estimation error regarding ATE and ITE on two semi-synthetic datasets. We report mean$\pm$std results. - means the method is not applicable. The best is bolded.}
    \label{Semi_result_table}
        \vspace{-.2cm}
\end{table}
\subsubsection{Is our method robust to different strengths of latent confounding?}
In table \ref{Dif_beta_Syn_result_table}, we compare our model with baselines on the fourth synthetic dataset with different strengths of confounding bias controlled by $\beta$.
The main observations are as follows.
With the strengths of latent confounding increasing, i.e., $\beta$ from $1$ to $5$, all methods perform worse, which is reasonable since a large confounding bias will lead to a significant imbalance of distribution between treated and control groups.
When the latent confounding is small, traditional methods yield a comparable performance, since the unconfoundedness assumption almost holds.
When the latent confounding is large enough, only our method yields accurate estimations in terms of ATE and ITE, 
which indicates that our method is robust to the latent confounding.
It is because our method can correctly recover the latent confounders $Z$,
and it also reveals the necessity of recovering latent confounders.

\section{Conclusion}
In this paper, we provide a practical solution to estimate the long-term individual causal effects in the presence of latent confounders via identifiable representation learning. Our proposed method takes advantage of the natural heterogeneity of data, e.g., data from multiple cities, to identify latent confounders and further estimate the long-term individual effect, which not only helps us avoid the idealized assumptions of the existing methods, but also renders our approach with theoretical guarantees of identifiability. Extensive experimental results verify the correctness of our theory and the effectiveness of our estimator.
\appendix

\section*{Acknowledgments}

This research was supported in part by National Science and Technology Major Project (2021ZD0111501), National Science Fund for Excellent Young Scholars (62122022), Natural Science Foundation of China (U24A20233, 62206064, 62206061, 62476163, 62406078),  Guangdong Basic and Applied Basic Research Foundation (2023B1515120020), and CCF-DiDi GAIA Collaborative Research Funds (CCF-DiDi GAIA 202311).

\bibliographystyle{named}
\bibliography{main}
\onecolumn 

\appendix 

\setcounter{proposition}{0}
\setcounter{theorem}{0}
\numberwithin{definition}{section}
\numberwithin{equation}{section}
\numberwithin{figure}{section}
\setcounter{figure}{0}
\setcounter{section}{0}

\section{Definition of Identifiability of Latent Variables }
\renewcommand{\theequation}{A.\arabic{equation}}

In this section, we provide the formal definition of the identifiability of latent variables discussed in Section 5.1. To be clear, we begin with our assumed causal graphs:

\begin{figure}[!h]
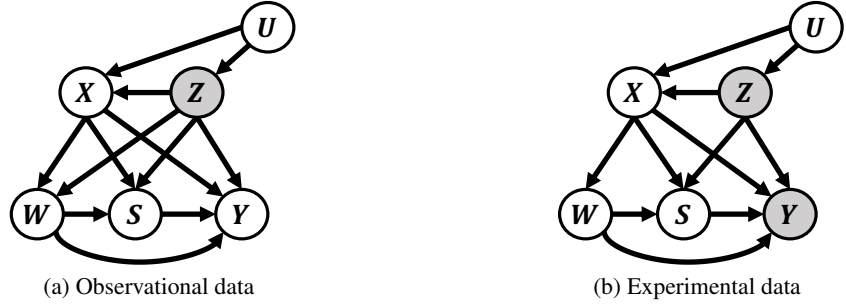
 
	\centering
	\subfloat[Observational data]{
		\includegraphics[width=0.22\textwidth]{./our_model_causal_graph_obs1.pdf} 
		\label{figure obs}
	}
	\hfil
	\subfloat[Experimental data]{
		\includegraphics[width=0.22\textwidth]{./our_model_causal_graph_exp1.pdf}
		\label{figure exp}
	}
	\caption{Two causal graphs in our setting. The white nodes denote observed variables and the grey denote unobserved variables. Fig. \ref{figure obs} is causal graphs of experimental in our setting. Fig. \ref{figure exp} is causal graph of observational data in our setting.}
	 \label{fig:causal graph} 
\end{figure}
As shown in Fig. \ref{fig:causal graph}, pre-treatment variable $X$, short-term outcome $S$ and treatment $W$ are generated from latent confounders(variables) $Z$ and auxiliary variable $U$ via a function $f_U$ which is shown in equation Eq. \eqref{ZU_gen_X}:
\begin{equation}
    X,S,W=f_U(Z)
    \label{ZU_gen_X}.
\end{equation}

\begin{definition}[Identifiability of Latent Variables $Z$]
    Let $X,S,W$ be observed
variables generated by the true latent causal process as shown in Fig.2. A learned generative model $\hat{f}_U$  is observationally equivalent
to $f_U$ if the model distribution $p_{\hat{f}_U}(X,S,W)$  matches the data distribution $p_{f_U}(X,S,W)$ everywhere. We say  latent causal processes are identifiable if observational
equivalence can lead to identifiability of the latent variables up to permutation $\Pi$ and component-wise
invertible transformation $\mathcal{T}$:
\begin{equation}
p_{f_U}(X,S,W)=p_{\hat{f}_U}(X,S,W)\Rightarrow \hat{f_U} =f_U \circ \Pi \circ \mathcal{T}.
\end{equation}
Once the process $f_U$ gets identified, the latent variables will be immediately identified up to
permutation and component-wise invertible transformation:
\begin{equation}
\hat{Z}=\hat{f}_U^{-1}(X,S,W)=(\mathcal{T}^{-1} \circ \Pi^{-1} \circ f^{-1})(X,S,W)=(\mathcal{T}^{-1} \circ \Pi^{-1})(f^{-1}_U(X,S,W))=(\mathcal{T}^{-1} \circ \Pi^{-1})Z.
\end{equation}
\end{definition}

\section{Discussion of The Conditions}
\renewcommand{\theequation}{B.\arabic{equation}}
In this section, we provide more implication of three conditions mentioned in Theorem 1. To be clear, we will divide the discussion into three parts, respectively elaborating on the implications of each of the three conditions.   \\
\textbf{Smooth and Positive Density
:} This conditions is common in the existing nonlinear ICA literature \citep{khemakhem2020variational,kong2022partial}. It denotes that the transition probabilities between latent confounders and  auxiliary variables are always positive. 
\\
\textbf{Conditional independence:} This condition is also standard in the identification of latent variables \citep{khemakhem2020variational,kong2022partial}. Intuitively, it means that there are no immediate relations among the latent variables given the auxiliary variables. 
For example, in a study on driver income in a ride-hailing platform, let $X$ represent observable driver characteristics, $W$ denote driver income, $S$ indicate short-term retention, and $Y$ represent long-term retention. The auxiliary variable $U$ is the city label, while the unobserved confounders $Z_1$ and $Z_2$ represent drivers' water and internet bill expenditures.
Since both expenditures reflect urban spending levels, $Z_1$ and $Z_2$ are naturally correlated. However, they may be independent when conditioned on $U$, as different bills could vary independently given a fixed urban spending level.
\\
\textbf{Linear independence:} Finally, this condition is also common in the identification of nonlinear ICA \citep{khemakhem2020variational,kong2022partial}. Mathematically, it means that the conditional distributions is a second-order derivative and the condition of the unique solution of the system of equations. Empirically, this condition is plausible due to the nature of the heterogeneity of data, e.g., data from 11 cities can ensure the identifiability of Z with up to 5 dimensions. When we obtain data from a sufficient number of cities, we can obtain richer information to make the second-order derivative of conditional distribution linear independence.

\section{Proof of The Latent Confounder Identifiability (Theorem 1)}
\renewcommand{\theequation}{C.\arabic{equation}}

Our proof structures partly follow \cite{khemakhem2020variational,kong2022partial,li2023subspace,cai2025disentangling,cai2025learning}.

\textbf{Proof sketch.} First, we construct an invertible transformation $h$ between the ground-truth $Z$ and estimated $\hat{Z}$. Next, we leverage the variance of different labels to construct a full-rank linear system, where the only solution of $\frac{\partial{Z_i}}{\partial{Z_k}} \frac{\partial{Z_i}}{\partial{Z_q}}$ and $\frac{\partial^2 Z_i}{\partial{\hat{Z}_k}\partial{\hat{Z}_q}}$ is zero, $i=1,...,d_z$ and $k,q\in [d_z]$ and $k\not= q$. Since the Jacobian of $h$ is invertible, there is exactly one non-zero element in each row of the Jacobian matrix of $h$, for each $Z_i,i\in\{1,...,d_z\}$, there exits a $h_i$ such that $Z_i=h_i(\hat{Z_i})$. Formal proofs are given as follow.

\begin{proof}
    
We begin with the matched marginal distribution $p_{X,S,W|U}$ to develop the relation between $Z$ and $\hat{Z}$ as follows: $\forall \  U \in \mathcal{U}$, 

\begin{equation}
     \begin{aligned}
       & p_{ \hat{X},\hat{S},\hat{W} | U } = p_{X,S,W | U} \\
       \Longleftrightarrow &  p_{ \hat{f}_U(\hat{Z}) | U } = p_{f_U(Z) | U} \\
       \Longleftrightarrow & 
       p_{f_U^{-1}\circ \hat{f}_U(\hat{Z})|U}|\mathbf{J}_{f_U^{-1}}| = p_{Z|U}|\mathbf{J}_{f_U^{-1}}| \\
       \Longleftrightarrow  & 
       p_{ h(\hat{Z}) |U } = p_{Z|U}, 
    \end{aligned}
    \label{proof A1 eq1}
\end{equation}
where $\hat{f}_{U}^{-1}: (X,S,W) \rightarrow Z$ denotes the estimated invertible inference function, and $h:= f^{-1}_U\circ \hat{f}_U$ is the transformation between ground-truth latent confounders and the estimated one. $|\mathbf{J}_{f^{-1}_U}|$ is the absolute value of Jacobian matrix determinant of $f^{-1}_U$, $|\mathbf{J}_{f^{-1}_U}| \not = 0$ and $h$ is invertible.

According to the conditional independence assumption in Theorem 1, we have: 

\begin{equation}
\begin{aligned}
   &p_{Z|U}(Z|U)=\prod \limits_{i=1}^{d_z} p_{Z_i|U}(Z_i); \\
   &p_{\hat{Z}|U}(\hat{Z}|U)=\prod \limits_{i=1}^{d_z} p_{\hat{Z_i}|U}(\hat{Z_i}).
\end{aligned}
\label{decompose z|u}
\end{equation}

We denote $q_i:= \log p_{Z_i|U}$ and $\hat{q}_i:= \log p_{\hat{Z_i}|U}$. Then, we can rewrite Eq. \eqref{decompose z|u} as follow:

\begin{equation}
\begin{aligned}
   &\log{p_{Z|U}(Z|U)}=\sum\limits_{i=1}^{d_z} q_{i}(Z_i,U); \\
   &\log{p_{\hat{Z}|U}(\hat{Z}|U)}=\sum\limits_{i=1}^{d_z} \hat{q}_{i}(\hat{Z_i},U).
\end{aligned}
\label{log decompose z|u }
\end{equation}
Combining Eq. \eqref{proof A1 eq1} and Eq. \eqref{log decompose z|u }, we can yield
\begin{equation}
     \begin{aligned}
       &p_{Z|U}=p_{ h(\hat{Z}) |U }=p_{ \hat{Z}|U }| \mathbf{J}_{h^{-1}} |\\
       \Longleftrightarrow 
       &\log p_{Z|U} =\log p_{ \hat{Z}|U }+\log | \mathbf{J}_{h^{-1}} | \\
       \Longleftrightarrow 
       &\sum\limits_{i=1}^{d_z} q_{i}(Z_i,U) + \log{| \mathbf{J}_{h}|}=\sum\limits_{i=1}^{d_z} \hat{q}_{i}(\hat{Z_i},U), 
    \end{aligned}
    \label{proof A1 eq2}
\end{equation}
where $\mathbf{J}_{h^{-1}}$ and $\mathbf{J}_{h}$ are the Jacobian matrix of the transformation associated with $h^{-1}$, $h$ respectively. 
For simplicity of expression, we introduce the following notation:
\begin{equation}
     \begin{aligned}
       &h^{'}_{i,(k)}:=\frac{\partial Z_i}{\partial{\hat{Z}_k}},  \ \ h^{''}_{i,(k,q)}:=\frac{\partial^2 Z_i}{\partial{\hat{Z}_k}\partial{\hat{Z}_q}};\\
       &\eta^{'}_{i}(Z_i,U):=\frac{\partial q_i(Z_i,U)}{\partial Z_i},\ \ \ \eta^{''}_{i}(Z_i,U):=\frac{\partial^2 q_i(Z_i,U)}{(\partial Z_i)^2}.
    \end{aligned}
    \label{simply notataion}
\end{equation}
Differentiating both sides of Eq. \eqref{proof A1 eq2} twice w.r.t. $\hat{Z}_k$ and $\hat{Z}_q$, where $k,q\in \{1,...,d_z\}$ and $k\not= q$ yields \begin{equation}
    \sum\limits_{i=1}^{d_z}{(\eta^{''}_{i}(Z_i,U) \cdot h^{'}_{i,(k)}h^{'}_{i,(q)}+\eta^{'}_{i}(Z_i,U) \cdot h^{''}_{i,(k,q)})} + \frac{\partial^2 \log{|\mathbf{J}_h|} }{\partial \hat{Z}_i\partial \hat{Z}_i}=0.
    \label{defferentiating both side twice}
\end{equation}
Therefore, for $U=u_0,...,u_{2d_z}$, we have $2n_z +1$ such equations. Subtracting each equation corresponding to $u_1,...,u_{2d_z}$ with the equation corresponding to $u_0$ results in $2d_z$ equations:
\begin{equation}
\begin{aligned}
    \sum\limits_{i=1}^{d_z}((\eta^{''}_{i}(Z_i,u_j)-\eta^{''}_{i}(Z_i,u_0)) \cdot h^{'}_{i,(k)}h^{'}_{i,(q)}+(\eta^{'}_{i}(Z_i,u_j)-\eta^{'}_{i}(Z_i,u_0)) \cdot h^{''}_{i,(k,q)}) = 0,
    \label{defferentiating both side twice Sub u0}
\end{aligned}
\end{equation}
where $j=1,...,2d_z$.

Under the linear independence condition in Theorem 1, the linear system is a $2d_z \times 2d_z$ full-rank system. Therefore the only solution is $h^{'}_{i,(k)}h^{'}_{i,(q)}=0$ and $h^{''}_{i,(k,q)}=0$ for $i=1,...,d_z$ and $k,q\in \{1,...,d_z\}$ and $k\not= q$.

Since $h(\cdot)$ is smooth over $\mathcal{Z}$, its Jacobian can written as follows:
\begin{equation}
\begin{aligned}
    {\mathbf{J}_{h}}_{(i,j)}=
    \begin{matrix}
        \frac{\partial Z_i}{\partial{\hat{Z}_j}},
    \end{matrix}
    \label{Jh}
\end{aligned}
\end{equation}
where $i, j\in \{1,...,d_z\}$. Note that $h^{'}_{i,(k)}h^{'}_{i,(q)}=0$ means that for each $i=1,..., d_z$, $h^{'}_{i,(k)}\not= 0$ for at most one element $k\in \{1,...,d_z\}$. So, there is only at most one non-zero element in each row in the Jacobian matrix  $\mathbf{J}_{h}$. Additionally, the $h(\cdot)$ is invertible which implies that there is exactly one non-zero element in each row of Jacobian matrix  $\mathbf{J}_{h}$. Therefore, latent confounders $Z$ is identified up to permutation and component-wise invertible transformations, i.e., for each ground-truth latent confounders $Z_i$, there exists a corresponding estimated latent confounder $\hat{Z_j}$ and an invertible function $h_i: \mathbb{R} \rightarrow \mathbb{R}$, 
such that $Z_i = h_i(\hat{Z_j})$.

\end{proof}

\section{Proof of The Long-term Individual Causal Effect Identication (Theorem 2)}
\renewcommand{\theequation}{D.\arabic{equation}}
\begin{proof} 

Based on Assumption 3 ($W\Vbar \{Y(w), S(w)\}| X, G=e$), the experimental data is unconfounded, and thus our module of Short-term Potential Outcome Estimation yields valid short-term potential outcomes of experimental data.
Under Assumption 4 ($G\Vbar \{Y(w), S(w)\}| X$), we have $ p(S(w)|X=x,G=e)=p(S(w)|X=x,G=o)$. Hence, our module of Short-term Potential Outcome Estimation also yields valid short-term potential outcomes of observational data.
With the correctly estimated potential outcome of $S(1),S(0)$ in observational data, i.e., $p(S(1),S(0)|G=o,X=x)$, suppose Theorem 1 holds, we obtain the estimated latent confounder $\hat Z = h^{-1}(Z)$.

Then, we can write down the expression of long-term individual causal effect (ITE) as follows:
\begin{equation}
\begin{aligned}
       \tau(x)
       &  =\mathbb{E}[Y(1)-Y(0)|X=x]
       \\
  &  =\mathbb{E}[Y(1)-Y(0)|X=x,G=o]\\
  & = \mathbb{E}[\mathbb{E}[Y(1)|W=1,X=x,Z=z,G=o]-\mathbb{E}[Y(0)|W=0,X=x,Z=z,G=o]]
  \\ &
  = \mathbb{E}[\mathbb{E}[Y|W=1,X=x_{i},Z=z,G=o]-\mathbb{E}[Y|W=0,X=x,Z=z,G=o]]
    \\ &
  = \mathbb{E}[\mathbb{E}[Y|W=1,X=x_{i},\hat{Z}=\hat{z},G=o]-\mathbb{E}[Y|W=0,X=x,\hat{Z}=\hat{z},G=o]],
\label{ITE_ID_proof_3}
\end{aligned}
\end{equation}
where the second equality is based on Assumption 4 ($G\Vbar \{Y(w), S(w)\}| X$), the third equality is based on Assumption 2 ($W\Vbar \{Y(w), S(w)\}|X, Z, G=o$), the fourth equality is based on Assumption 1 ($Y=Y(w)$ if $W=w$), and the last equality holds due to the invertibility of $h$.

\end{proof}

\section{Derivation of ELBO}
\renewcommand{\theequation}{E.\arabic{equation}}

In this section, we show the evident lower bound. We first factorize the conditional distribution according to the Bayes theorem.
\begin{equation}  
  \begin{aligned}
    p_{\theta}(X,W,S|U)=p_{\theta}(X,W,S(w)|U)=\frac{p_{\theta}(X,W,S(w),Z|U)}{p_{\theta}(Z|X,W,S(w),U)},
   \label{ELBO step 1} 
  \end{aligned}
\end{equation}
where the first equality is based on Assumption 1. A1. The second equality  can be rewritten as:
\begin{equation}  
  \begin{aligned}
   \log p_{\theta}(X,W,S(w)|U) =\log p_{\theta}(X,W,S(w),Z|U)-\log p_{\theta}(Z|X,W,S(w),U).
   \label{ELBO step 2} 
  \end{aligned}
\end{equation}

Further rewrite Eq. \eqref{ELBO step 2}, we have
\begin{align}
   &\int_{q_{\phi}(Z|X,W,S(1),S(0),U)} \left[ \log p_{\theta}(X,W,S(w)|U) \right] 
   q_{\phi}(Z|X,W,S(1),S(0),U) \mathrm{d}Z \nonumber \\
   &= \int_{q_{\phi}(Z|X,W,S(1),S(0),U)} \left[ \log p_{\theta}(X,W,S(w),Z|U) - \log p_{\theta}(Z|X,W,S(w),U) \right] 
   q_{\phi}(Z|X,W,S(1),S(0),U)  \mathrm{d}Z, \label{ELBO step 3} \\[10pt]
   &\log p_{\theta}(X,W,S(w)|U) \nonumber \\
   &=  \int_{q_{\phi}(Z|X,W,S(1),S(0),U)} q_{\phi}(Z|X,W,S(1),S(0),U) \left[ 
    \log  \frac{p_{\theta}(X,W,S(w),Z|U)}{q_{\phi}(Z|X,W,S(1),S(0),U)}  \right] \mathrm{d}Z \nonumber \\
    &\quad + \mathrm{KL}\left[q_{\phi}(Z|X,W,S(1),S(0),U) || p_{\theta}(Z|X,W,S(w),U) \right]. \label{ELBO step 4}
\end{align}
The first term on the right side of Eq. \eqref{ELBO step 4} is ELBO.

\begin{equation}  
  \begin{aligned}
   \mathrm{ELBO} = & \int_{q_{\phi}(Z|X,W,S(1),S(0),U)} q_{\phi}(Z|X,W,S(1),S(0),U) \left[ 
    \log  \frac{p_{\theta}(X,W,S(w),Z|U)}{q_{\phi}(Z|X,W,S(1),S(0),U)}  \right] \mathrm{d}Z \\
    =  &\int_{q_{\phi}(Z|X,W,S(1),S(0),U)} q_{\phi}(Z|X,W,S(1),S(0),U)\left[ 
    \log p_{\theta}(Z|U) + \log p_{\theta}(X|Z,U) + \log p_{\theta}(W|X,Z,U) \right. \\
    &\quad \left. + \log p_{\theta}(S(w)|W,Z,X,U) - \log q_{\phi}(Z|X,W,S(1),S(0),U) \right] \mathrm{d}Z \\
    =  &\int_{q_{\phi}(Z|X,W,S(1),S(0),U)} q_{\phi}(Z|X,W,S(1),S(0),U) \left[ 
     \log p_{\theta}(X|Z,U) +\log p_{\theta}(W|X,Z)+\log p_{\theta}(S|W,Z,X)\right] \mathrm{d}Z \\
     &- \mathrm{KL}\left[ q_{\phi}(Z|X,W,S(1),S(0),U) || p_{\theta}(Z|U)  \right]. \\
   \label{ELBO step 5} 
  \end{aligned}
\end{equation}
From the above ELBO, we can obtain encoder $q_{\phi}(Z|X,W,S(1),S(0),U)$ as well as decoder $p_{\theta}(X|Z,U)$, $p_{\theta}(W|X,Z)$, $p_{\theta}(S|W,Z,X)$.

\section{Experimental details}
\renewcommand{\theequation}{F.\arabic{equation}}

\subsection{Reproducibility}

Our model was implemented using Python 3.7.16 and PyTorch 1.10.2 framework. All experiments are run on Windows11 22631.4037 (64bit) with CPU: 13th Gen Intel(R) Core(TM) i9-13900KF 3.00 GHz; GPU: NVIDIA GeForce RTX4090; Memory:128 GB DDR4 RAM.  We show the hyperparameter space in Table \ref{tb: param}.

\begin{table}[h] 
\centering
\begin{tabular}{@{}lp{5cm}@{}}
\toprule
 \textbf{Hyperparameter} & \textbf{Tuning Range} \\ 
\midrule
Hidden units of each layer & $32,64,128$ (Synthetic data 1-5) \\
&  $64,128,256$ (IHDP)  \\
& $64,256,512$ (TWINS)      \\
Learning rate & $0.001,0.0001$ \\
 Batch size & 32, 50, 100 \\
 Epoch Number & 200,300,400 \\
 Optimizer & Adam \\
\midrule
\end{tabular}
\caption{Hyperparameters Space.} \label{tb: param}
\end{table}

\noindent\textbf{Baseline Methods:} We clarify the baseline methods as follows:
\begin{itemize}[leftmargin=2em]
  \item \textbf{CEVAE \citep{louizos2017causal}:} CEVAE recovers the latent confounder $Z$ by treating the pre-treatment variable $X$ as proxies via VAE method, and then controls $Z$ to estimate the causal effect. This method does not control latent confounders $X$ when estimating the long-term outcome. As a result, in our scenario, it fails to block the back-door path from treatment to the long-term potential outcome, making it unsuitable for long-term scenarios. Therefore, we modify the long-term estimator from $\mathbb{E}[Y|W, Z]$ to $\mathbb{E}[Y|W, X, Z]$.  We use the code available at \url{https://github.com/pyro-ppl/pyro/blob/dev/examples/contrib/cevae}.
  
  \item \textbf{TEDVAE \citep{zhang2021treatment}:}  TEDVAE is a follow-up work of CEVAE, which divides the latent confounders $Z$ into three factors, including $Z_w$ affecting $W$, $Z_y$ affecting $Y$ and $Z_c$ affecting $W$ and $Y$.
  We use the code available at \url{https://github.com/WeijiaZhang24/TEDVAE}. 
  \item \textbf{LTEE \citep{cheng2021long}:} LTEE employs a dual-headed RNN to separately capture the relationships of time-dependent surrogates in the treatment and control groups, and then estimates the CATE using the predicted long-term outcomes. We use the code available at \url{https://github.com/GitHubLuCheng/LTEE}. 
  \item \textbf{S-Learner and T-Learner \citep{kunzel2019metalearners}:} S-Learner and T-Learner estimate long-term outcomes using a single estimator ($\mathbb{E}[Y|X, W]$) and two estimators ($\mathbb{E}[Y|X, W=0]$ and $\mathbb{E}[Y|X, W=1]$), respectively. In our experiments, these estimators are implemented using MLPs.
  \item \textbf{Imputation and Weighting \citep{athey2020combining}:} Both are based on latent unconfoundedness assumption. The Imputation approach fits a regression of the long-term outcome in the observational sample, imputes the long-term outcome for the experimental sample using the fitted regression, and then adjusts for covariates to estimate the long-term average treatment effect in observational data. In our experiments, all the estimators in this method are implemented using MLPs. 
  The Weighting approach estimates the probability densities in both the observational and experimental samples, constructs weights to adjust for differences between them, and uses these weights and probability densities to estimate the long-term average treatment effect. In our experiments, we use Gaussian kernel functions to estimate these conditional joint probability densities.   
  \item \textbf{Equi-naive and IF-base \citep{ghassami2022combining}:} Both are based on conditional additive equi-confounding bias assumption. The Equi-naive method first estimates the short-term confounding bias (equal to the long-term one) by combining observational and experimental data, and then constructs the final estimator using the estimated confounding bias.
  The IF-based estimator is implemented according to the derived efficient influence curve (see Section 7.1 in \cite{ghassami2022combining}).
  In our experiments, all the estimators in these two methods are implemented using MLPs.
\end{itemize}
Our code will be available upon acceptance.



\subsection{Data Generation Process}

The first synthetic data (\textbf{Synthetic 1}) corresponds to the assumption of our method, the second synthetic data (\textbf{Synthetic 2}) corresponds to the assumption of Athey et al.(2020), the third synthetic data (\textbf{Synthetic 3}) corresponds to the assumption of Ghassami et al.(2022). We use these datasets to verify the generalizability of our method. The fourth synthetic dataset (\textbf{Synthetic 4}) is used to analyze performance in terms of different strengths of confounding bias.

\begin{figure}[!ht] 
	\centering
	\subfloat[\textbf{Synthetic 1\& 4\& 5}]{
		\includegraphics[width=0.20\textwidth]{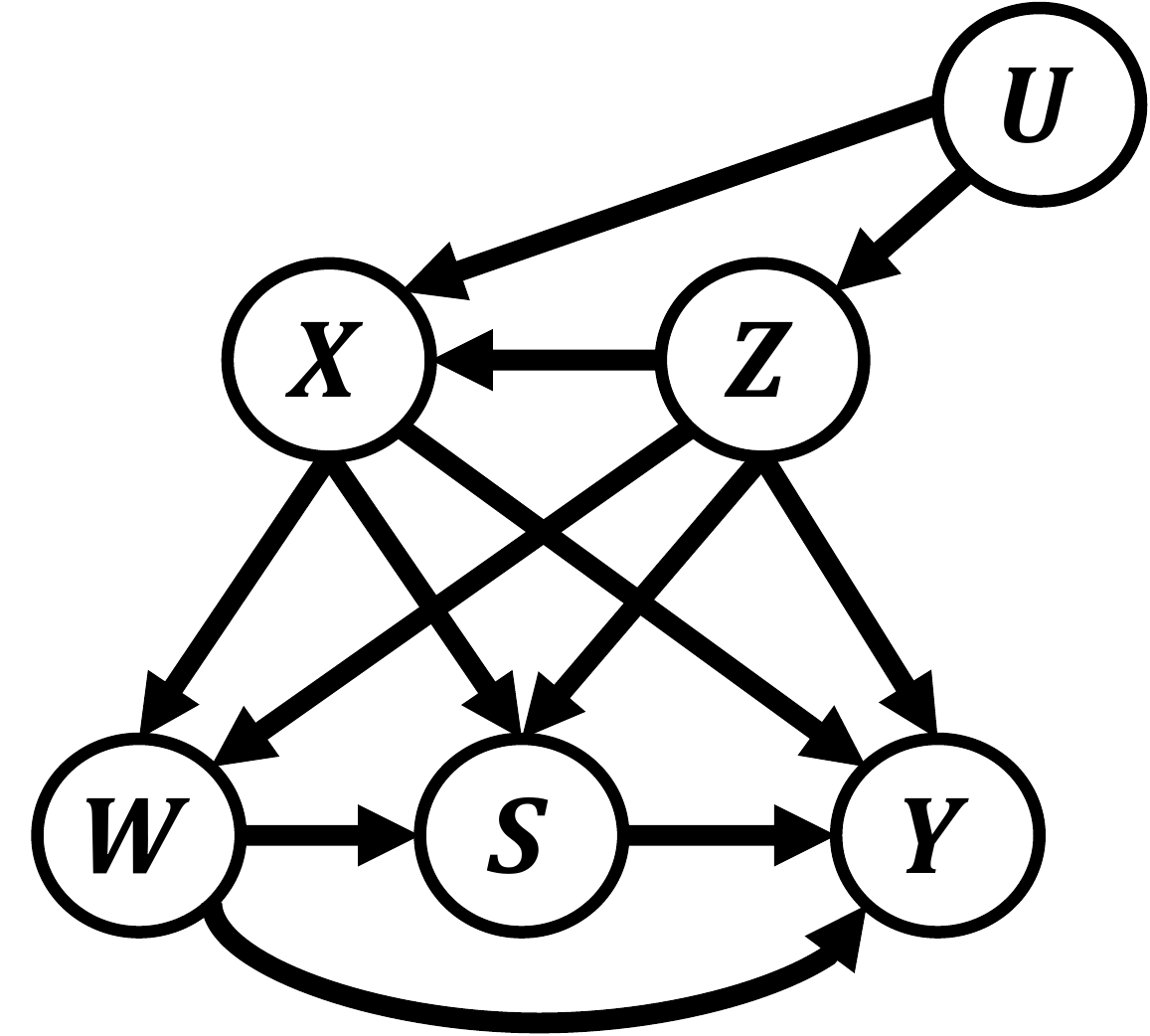} 
		\label{syn1 graph}
	}\hspace{4em}
	\subfloat[\textbf{Synthetic 2}]{
		\includegraphics[width=0.20\textwidth]{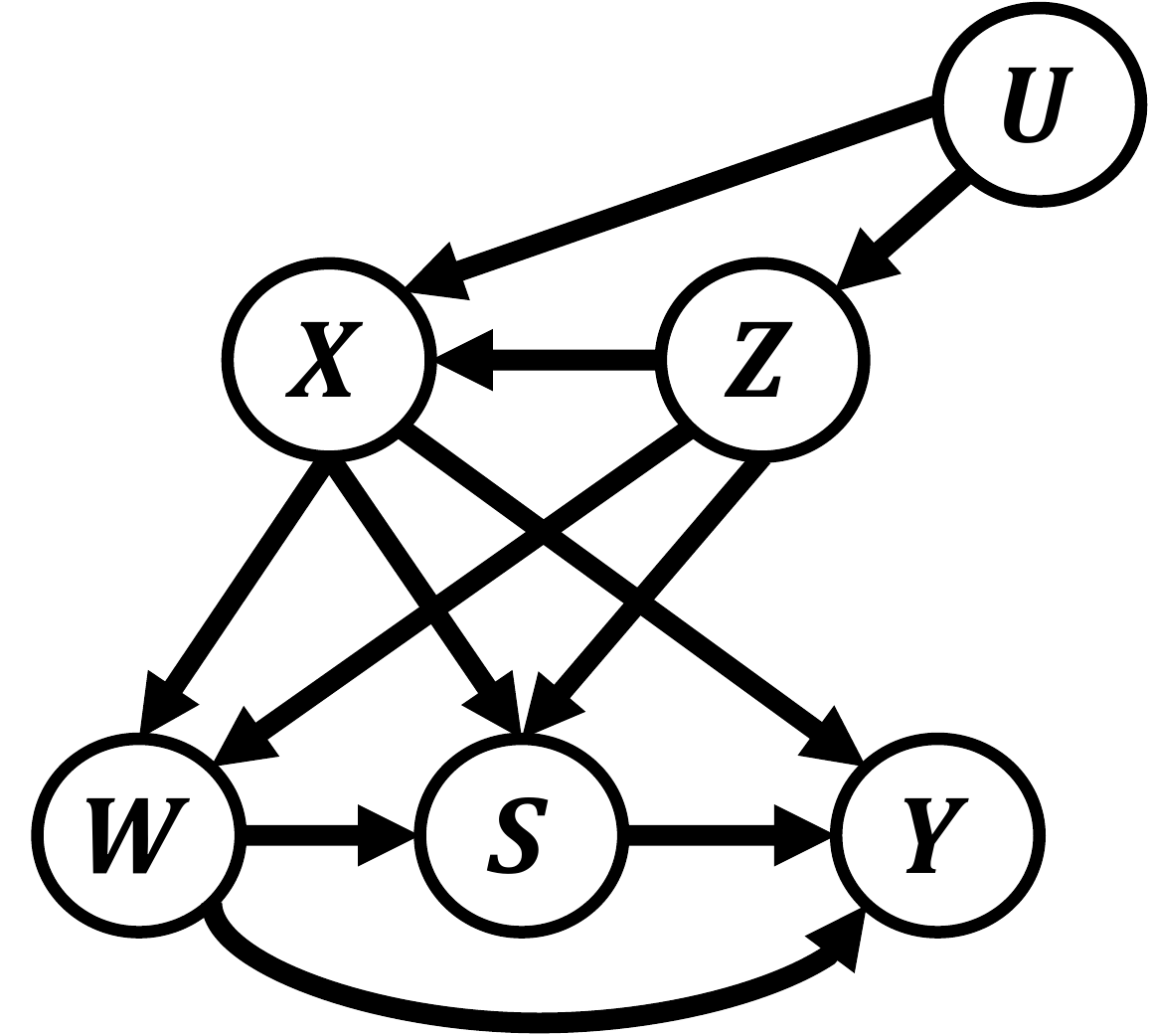}
		\label{syn2 graph}
	}\hspace{4em}
   \subfloat[\textbf{Synthetic 3}]{
		\includegraphics[width=0.20\textwidth]{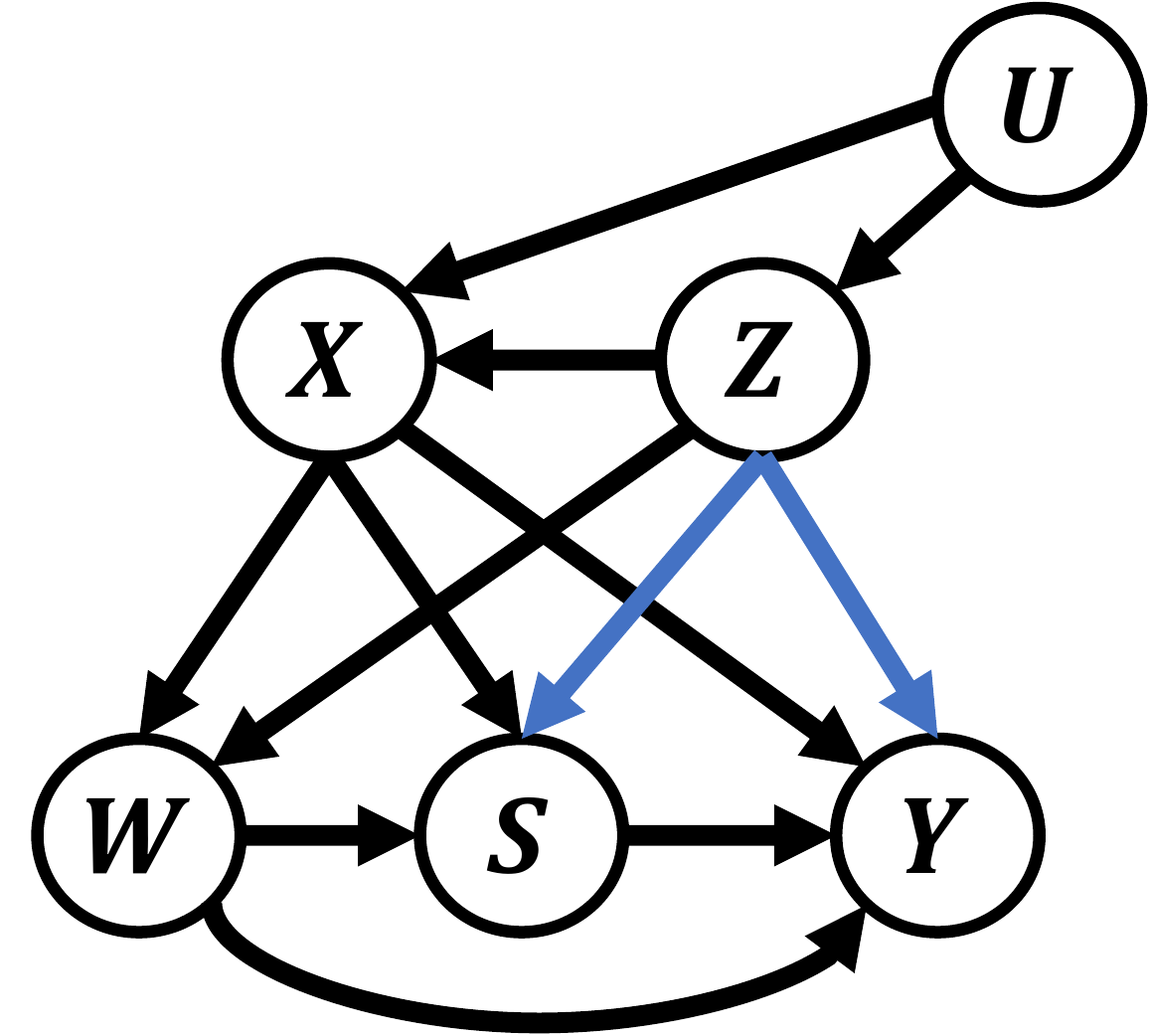}
		\label{syn3 graph}
	}
	\caption{The variables are defined as follows: $X$ represents the pre-treatment variable; $Y$ denotes the long-term outcome; $Z$ is the latent confounders; $S$ is the short-term outcome; $U$ is the auxiliary variable; and $W$ represents the treatment. the blue arrow in Fig. \ref{syn3 graph} shows the same confounding bias.
    }
     \label{fig: syn1-syn3} 
\end{figure}

Since three of the synthetic datasets (\textbf{Synthetic 1}, \textbf{Synthetic 2}, \textbf{Synthetic 3}) share a common data generation process involving the variables $\{X, W, Z, U\}$, we will first outline the shared aspects of these datasets. Subsequently, we will detail the unique characteristics of each dataset. Our data generation process is partly motivated by \cite{yang2024estimating}.

To simplify notations, we denote $X^e$ as the covariate of experimental data, i.e. $X|G=e$, and similarly for $X^o,Z^e,Z^o,W^e,W^o,U^e$ and $U^o$. For each unit in observational data, we first generate one-dimensional auxiliary variable, i.e., sample $U^o$ uniformly from $\{0,1,2,3,4\}$. Subsequently we generate two-dimensional latent confounders, i.e., $Z^o_1 \sim \mathcal{N}(- U^o, 1)$ and $Z^o_2 \sim \mathcal{N}(2 U^o, 1)$. Then, we generate two-dimensional observed pre-treatment variables $X^o$ based on $Z^o$. i.e., $X^o_1 \sim \mathcal{N}(Z^o_1+0.5U^o, 1)$ and  $X^o_2 \sim \mathcal{N}(0.5Z^o_2+U^o, 1)$. Similarly, we generate units in experimental data as sample $U^e$ uniformly from $\{0,1,2,3,4 \}$, $Z^e_1 \sim \mathcal{N}(- U^e, 1)$, $Z^e_2 \sim \mathcal{N}(2 U^e, 1)$, $X^e_1 \sim \mathcal{N}(Z^e_1+0.5U^e, 1)$, $X^e_2 \sim \mathcal{N}(0.5Z^e_2+U^e, 1)$. Then we generate treatment for these two groups of datasets: 
\begin{equation}  
    \begin{aligned}
          W^o=&\text{Bern}\left(\frac{1}{1+e^{
          -\left(
          \frac{X^o_1+X^o_2+3Z^o_1+3Z^o_2}{4}
          \right)}}
          \right),
    \end{aligned}
\end{equation}

\begin{equation}  
    \begin{aligned}
          W^e=&\text{Bern}\left(\frac{1}{1+e^{
          -\left(
          \frac{X^e_1+X^e_2}{2}
          \right)}}
          \right).
    \end{aligned}
\end{equation}
In the first synthetic data (\textbf{Synthetic 1}), we generate one-dimensional short-term outcome $S$ and one-dimensional long-term outcome $Y$ as follows:

\begin{equation}  
    \begin{aligned}
          S=&3W+(2+W)\frac{X_1+X_2}{2}+\frac{Z_1+Z_2}{2}+\epsilon_{S},
    \end{aligned}
\end{equation}
\begin{equation}  
    \begin{aligned}
          Y=&4W+(1+W)\frac{X_1+X_2}{2}+S+3\frac{Z_1+Z_2}{2}+\epsilon_{Y}.
    \end{aligned}
\end{equation}
where $\epsilon_{S},\epsilon_{Y}$ are noise terms sampled from Gaussian distributions.
 
 In the second synthetic data (\textbf{Synthetic 2}), we generate one-dimensional short-term outcome $S$ and one-dimensional long-term outcome $Y$ as follows:
\begin{equation}  
    \begin{aligned}
          S=&3W+(2+W)\frac{X_1+X_2}{2}+\frac{Z_1+Z_2}{2}+\epsilon_{S},
    \end{aligned}
\end{equation}

\begin{equation}  
    \begin{aligned}
          Y=&4W+(1+W)\frac{X_1+X_2}{2}+S+\epsilon_{Y}.
    \end{aligned}
\end{equation}

 In the third synthetic data (\textbf{Synthetic 3}), we generate one-dimensional short-term outcome $S$ and one-dimensional long-term outcome $Y$ as follows:
  \begin{equation}  
    \begin{aligned}
          S=&1+W+(1.5+3W)(X_1+X_2)+Z_1+Z_2+\epsilon_{S},
    \end{aligned}
\end{equation}

\begin{equation}  
    \begin{aligned}
          Y=&2+3W+(2+6W)(X_1+X_2)+2(Z_1+Z_2)-S+\epsilon_{Y}.
    \end{aligned}
\end{equation}


Next, we present the data generation process for the fourth synthetic dataset (\textbf{Synthetic 4}). For each unit in observational data, we first generate one-dimensional auxiliary variable, i.e., sample $U^o$ uniformly from $\{0,1,2,3,4\}$. Subsequently we generate two-dimensional latent confounders, i.e., $Z^o_1 \sim \mathcal{N}(- U^o, 1)$ and $Z^o_2 \sim \mathcal{N}(2 U^o, 1)$. Then, we generate two-dimensional observed pre-treatment variables $X^o$ based on $Z^o$. i.e., $X^o_1 \sim \mathcal{N}(Z^o_1 + 0.5U^o, 1)$ and  $X^o_2 \sim \mathcal{N}(0.5Z^o_2 + U^o, 1)$. Similarly, we generate each unit in experimental data as sample $U^e$ uniformly from $\{0,1,2,3,4 \}$, $Z^e_1 \sim \mathcal{N}(- U^e, 1)$, $Z^e_2 \sim \mathcal{N}(2 U^e, 1)$, $X^e_1 \sim \mathcal{N}(Z^e_1 + 0.5U^e, 1)$, $X^e_2 \sim \mathcal{N}(0.5Z^e_2 + U^e, 1)$. Then we generate treatment for these two groups of datasets: 
\begin{equation}  
    \begin{aligned}
          W^o=&\textbf{Bern}\left(\frac{1}{1+e^{
          -\left(\frac{X^o_1+X^o_2+\beta (Z^o_1+Z^o_2)}{4}
          \right)}}
          \right),
    \end{aligned}
\end{equation}

\begin{equation}  
    \begin{aligned}
          W^e=&\textbf{Bern}\left(\frac{1}{1+e^{
          -\left(\frac{X^e_1+X^e_2}{2}
          \right)}}
          \right).
    \end{aligned}
\end{equation}\\
Where $\beta \in \{1,1.5,3,4.5,5\}$ denote different strengths of latent confounding bias. Next, we generate one-dimensional short-term outcome $S$ and one-dimensional long-term outcome $Y$ as follows:
\begin{equation}  
    \begin{aligned}
          S=&3W+(2+W)\frac{X_1+X_2}{2}+\frac{Z_1+Z_2}{2}+\epsilon_{S},
    \end{aligned}
\end{equation}

\begin{equation}  
    \begin{aligned}
          Y=&4W+(1+W)\frac{X_1+X_2}{2}+S+ \beta \frac{Z_1+Z_2}{2}+\epsilon_{Y}.
    \end{aligned}
\end{equation}

where $\epsilon_{S},\epsilon_{Y}$ are noise terms sampled from Gaussian distributions. Note that \textbf{Synthetic 1} is a special case of \textbf{Synthetic 4}. When the $\beta$ of \textbf{Synthetic 4} is equal to 3, \textbf{Synthetic 4} is equivalent to \textbf{Synthetic 1}.

Next, we present the data generation process for the fifth synthetic dataset (\textbf{Synthetic 5}). For each unit in observational data, we first generate one-dimensional auxiliary variable, i.e., sample $U^o$ uniformly from $\{0,1,2,3,4\}$. Subsequently we generate two-dimensional latent confounders, i.e., $Z^o_1 \sim \mathcal{N}(- U^o, 2.5)$ and $Z^o_2 \sim \mathcal{N}(1.5 U^o, 2.5)$. Then, we generate two-dimensional observed pre-treatment variables $X^o$ based on $Z^o$. i.e., $X^o_1 \sim \mathcal{N}(0.8Z^o_1 + 0.4Z^o_2 - 1.8U^o, 1)$ and  $X^o_2 \sim \mathcal{N}(0.4Z^o_1 + 0.4Z^o_2 + 1.8U^o, 1)$. Similarly, we generate each unit in experimental data as sample $U^e$ uniformly from $\{0,1,2,3,4\}$, $Z^e_1 \sim \mathcal{N}(- U^e, 2.5)$, $Z^e_2 \sim \mathcal{N}(1.5 U^e, 2.5)$, $X^e_1 \sim \mathcal{N}(0.8Z^e_1 + 0.4Z^e_2 - 1.8U^e, 1)$, $X^e_2 \sim \mathcal{N}(0.4Z^e_1 + 0.4Z^e_2 + 1.8U^e, 1)$. Then we generate treatment for these two groups of datasets: 
\begin{equation}  
    \begin{aligned}
          W^o=&\textbf{Bern}\left(\frac{1}{1+e^{
          -\left(\frac{X^o_1+X^o_2+Z^o_1+Z^o_2}{4}
          \right)}}
          \right),
    \end{aligned}
\end{equation}

\begin{equation}  
    \begin{aligned}
          W^e=&\textbf{Bern}\left(\frac{1}{1+e^{
          -\left(\frac{X^e_1+X^e_2}{2}
          \right)}}
          \right).
    \end{aligned}
\end{equation}\\
Next, we generate one-dimensional short-term outcome $S$ and one-dimensional long-term outcome $Y$ as follows:
\begin{equation}  
    \begin{aligned}
          S=&3W+(2+W)\frac{X_1+X_2}{2}+\frac{Z_1+Z_2}{2}+\epsilon_{S},
    \end{aligned}
\end{equation}

\begin{equation}  
    \begin{aligned}
          Y=&4W+(1+W)\frac{X_1+X_2}{2}+S+ \frac{Z_1+Z_2}{2}+\epsilon_{Y}.
    \end{aligned}
\end{equation}
where $\epsilon_{S},\epsilon_{Y}$ are noise terms sampled from Gaussian distributions.

In semi-synthetic datasets \textbf{IHDP} and \textbf{TWINS}, we reuse their covariates as the variables $\{X,  Z, U\}$ and generate the rest of the variables $\{W,  S, Y\}$. We will provide a detailed explanation of how to generate semi-synthetic data using \textbf{TWINS} and \textbf{IHDP}, respectively.

In semi-synthetic datasets \textbf{IHDP}, we use the three covariates columns $col_{10}, col_{11}, col_{12}$, which represent the binary variables for mother education level, and the seven columns $col_{19}, col_{20}, col_{21}, col_{22}, col_{23}$ $, col_{24}, col_{25}$, which represent the binary variables for regions of residence, to combine and form our auxiliary variable $U$ as follow:
\begin{equation}  
    \begin{aligned}
          U=&(3col_{10}+2col_{11}+col_{12})\left(\sum^{6}_{i=0} col_{(25-i)} 2^i \right).
    \end{aligned}
\end{equation}
Then, we choose five covariates columns to be five dimensions of latent confounders $Z$, each covariate representing "cigarettes", "alcohol", "drugs", "work during pregnancy" and " prenatal care". Next, we choose seven covariates columns to be seven dimensions of pre-treatment covariates $X$, each covariate representing "birth weight", "birth head circumference", "preterm birth", "birth order", "neonatal health index", "sex of the infant", "twin status". Then, we randomly partition ${X,Z,U}$ into observational $(G = o)$ and experimental
$(G = e)$ subsets ${X^o,Z^o,U^o,X^e,Z^e,U^e}$ at a 6:4 ratio.

The reason we selected these variables as 
$X$, $Z$, and $U$ is that $U$ has 384 potential values, while the selected 
$Z$ has only 5 dimensions. This satisfies the conditions of Theorem 1 (Linear independence). Moreover, a mother's education level and regions of residence, represented by $U$, can directly influence her personal lifestyle, represented by $Z$. Further, these personal lifestyles $Z$ can directly influence the features of the infant, represented by $X$.

Subsequently, we generate treatment $W^o$ and $W^e$ for these two subset  with coefficients $\alpha^{xw}_j \sim \mathcal{U}(-0.5, 0.5)$,$\alpha^{zw}_j \sim \mathcal{U}(-0.5, 0.5)$ as follow:
\begin{equation}  
    \begin{aligned}
    W_{temp}=&\left(\frac{0.25\mathbb{I}(G=o)\sum_{j=1}^5 \alpha^{zw}_j Z_j + \sum_{j=1}^7 \alpha^{xw}_j X_j}
         {\mathbb{I}(G=o)\frac{\sum_{j=1}^5 Z_j}{5} + \frac{\sum_{j=1}^7 X_j}{7}}
         \right),\\
         W=&\textbf{Bern}\left(\frac{1}{1+e^{W_{temp}}}
         \right).
         \end{aligned}
\end{equation}
Then, we generate short-term outcome and long-term outcomes with coefficients $\alpha^y_j \sim \mathcal{U}(0.5,1)$ as follows: 
\begin{equation}  
    \begin{aligned}
         Y_{(t)}=&\sum\limits_{j=1}^{4} \alpha^y_{j+5} X^2_j +(t+5)W\left(\sum\limits_{j=5}^{7} \alpha^y_{j+5} X_j \right)+\frac{\beta_Z t}{5}\left(\sum\limits_{j=1}^{5} \alpha^y_j cos Z_j \right)+0.25\overline{Y}_{(1:t-1)} + \epsilon_Y,
        \label{Synethic DGP}
    \end{aligned}
\end{equation}
where $\epsilon_{Y}$ is noise term sampled from Gaussian distributions, $Y_{(t)}$ denote the one-dimensional outcome at time point $t$,   $\overline{Y}_{1:t-1}$ denote the mean of $Y_1$ to $Y_{t-1}$. We denote short-term outcomes as $S = \{Y_1,Y_2,...,Y_7\}$ and long-term outcome as $Y=Y_{14}$.

In semi-synthetic datasets \textbf{TWINS}, we choose the covariate column "dmeduc", which represents the mother's education level, to be the auxiliary variable $U$. Then, we choose five covariates columns to be five dimensions of latent confounders $Z$, each covariate representing "drink", "cigar", "residential status", "adequacy of prenatal care utilization" and "detailed marital status". Next, we choose ten covariates columns to be ten dimensions of pre-treatment variables $X$, each covariate representing "number of prenatal visits", "cardiac disease", "total order of delivery", "diabetes", "chronic hypertension", "pregnancy-induced hypertension", "weight gain during pregnancy", "mother's age at delivery", "gestational age". Then, we randomly partition ${X,Z,U}$ into observational $(G = o)$ and experimental
$(G = e)$ subsets ${X^o,Z^o,U^o,X^e,Z^e,U^e}$ at a 6:4 ratio.

The reason we selected these variables as 
$X$, $Z$, and $U$ is that $U$ has 18 potential values, while the selected 
$Z$ has only 5 dimensions. This satisfies the conditions of Theorem 1 (Linear independence). Moreover, a mother's education level, represented by $U$, can directly influence her current living situation, represented by $Z$. Further, these personal current living situations $Z$ can directly influence the indicators during pregnancy, represented by $X$.

Subsequently, we generate treatment $W^o$ and $W^e$ for these two subset  with coefficients $\alpha^{xw}_j \sim \mathcal{U}(-0.5, 0.5)$,$\alpha^{zw}_j \sim \mathcal{U}(-0.5, 0.5)$ as follow:
\begin{equation}  
    \begin{aligned}
    W_{temp}=&\left(\frac{0.5\mathbb{I}(G=o)\sum_{j=1}^5 \alpha^{zw}_j Z_j + \sum_{j=1}^{10} \alpha^{xw}_j X_j}
         {\mathbb{I}(G=o)\frac{\sum_{j=1}^5 Z_j}{5} + \frac{\sum_{j=1}^{10} X_j}{10}}
         \right),\\
         W=&\textbf{Bern}\left(\frac{1}{1+e^{
         W_{temp}
         }}
         \right).
         \end{aligned}
\end{equation}
Then, we generate short-term outcome and long-term outcomes with coefficients $\alpha^y_j \sim \mathcal{U}(0.5,1)$ as follows: 
\begin{equation}  
    \begin{aligned}
         Y_{(t)}=&\sum\limits_{j=1}^{5} \alpha^y_{j+5} X^2_j +(t+5)W\left(\sum\limits_{j=6}^{10} \alpha^y_{j+5} X_j \right)+\frac{\beta_Z t}{5}\left(\sum\limits_{j=1}^{5} \alpha^y_j cos Z_j \right)+0.25\overline{Y}_{(1:t-1)} + \epsilon_Y,
    \end{aligned}
\end{equation}
where $\epsilon_{Y}$ is noise term sampled from Gaussian distributions, 
 $\overline{Y}_{(1:t-1)}$ denote the mean of $Y_1$ to $Y_{t-1}$. We denote short-term outcomes as $S = \{Y_1,Y_2,...,Y_7\}$ and long-term outcome as $Y=Y_{14}$.

\section{Additional Experiments}
\renewcommand{\theequation}{G.\arabic{equation}}
In this section, we will present the supplementary experiments to answer the following questions:
\begin{itemize}[leftmargin= 8px]
    \item[] 1. \textbf{If the Linear Independence Assumption is violated, is it still possible for our model to identify latent confounders?}
    \item[] 2. \textbf{How much experimental data is required for training our model?}
\end{itemize}

\subsection{Verification of the Linear Independence Assumption}
In the Synthetic 5 dataset, the dimension of the latent confounder $Z$ is $d_z=2$.  The \textbf{Linear Independence} assumption requires that the label variable $U$ has at least $2d_z+1$ distinct values. To test the \textbf{Linear Independence} assumption, we modified the generation method of $U$ while keeping all other aspects of the data generation process (DGP) of the Synthetic 5 dataset unchanged. 
Specifically, we draw sample $U^o,U^e$ uniformly from $\{0,1,..,d_u \}$ where $d_u \in \{2,3,4,5,6\}$, generating five new datasets. Here, $d_u+1$ represents the number of distinct values of $U$. 
Then, we applied our method to these datasets for experiments. 
As shown in Table 2, MCC drops sharply when fewer than 5 values are used, the minimum required by the assumption. 
\begin{table}[h]
\centering
\begin{tabular}{|c|c|c|c|c|c|}
\hline
$d_u+1$ & 3 & 4 & 5 & 6 & 7 \\\hline
MCC & 0.42 & 0.52 & 0.80 & 0.90 & 0.91  \\
\hline
\end{tabular}
\caption{Result on five datasets with different numbers of $d_u+1$.}\label{tab:tab2}
\end{table}

\subsection{Experimental Sample Size Requirement}
To answer how much experimental data is required for training our model, we conduct experiments by applying our method to the Synthetic 1 dataset with varying experimental sample sizes. As shown in Table 3, while the experimental sample sizes increase, PEHE decreases rapidly. We found that 500 is a decent choice for the sample size (here the observational sample size is fixed at 2000).

\begin{table}[h]
\centering
\begin{tabular}{|c|c|c|c|c|c|c|}
\hline
exp size & 50 & 150 & 250 & 500 & 2000 & 10000 \\\hline
PEHE & 4.98 & 3.99 & 3.49 & 3.17 &  3.22 & 2.81 \\
\hline
\end{tabular}
\caption{Result of experimental dataset with different training sizes on Synthetic 1 dataset. }\label{tab:tab3}
\end{table}


\end{document}